\documentclass{article} 
\usepackage{iclr2025,times}

\usepackage{hyperref}

\usepackage{algorithm}
\usepackage{algorithmic}
\usepackage{forloop}

\usepackage{multirow}
\usepackage{graphicx}
\usepackage{subfigure}
\usepackage{subcaption}
\usepackage{placeins}
\usepackage[utf8]{inputenc} 
\usepackage{fancyhdr}       
\usepackage[T1]{fontenc}    
\usepackage{url}            
\usepackage{booktabs}       
\usepackage{amsfonts}       
\usepackage{nicefrac}       
\usepackage{microtype}      

\usepackage{amsmath}
\usepackage{amssymb}
\usepackage{mathtools}
\usepackage{amsthm}
\usepackage{tikz}

\usepackage[capitalize,noabbrev]{cleveref}

\newcommand{\newcommenter}[3]{%
    \expandafter\newcommand\csname #1\endcsname[1]{%
        \textcolor{#3}{[\textbf{#2}: ##1]}%
    }
}
\newcommenter{rahul}{Rahul}{green} 
\newcommenter{arka}{Arka}{red}

\theoremstyle{plain}
\newtheorem{theorem}{Theorem}[section]

\theoremstyle{definition}
\newtheorem{definition}[theorem]{Definition}

\theoremstyle{remark}

\usepackage{xcolor}
\definecolor{mydarkblue}{rgb}{0,0.08,0.45}
\hypersetup{
colorlinks=true,
linkcolor=mydarkblue,
citecolor=mydarkblue,
filecolor=mydarkblue,
urlcolor=mydarkblue,
}

\newcommand{\newmethod}{Cascade}

\usepackage{amsmath,amsfonts,bm}

\def\eqref#1{equation~\ref{#1}}

\def\1{\bm{1}}

\DeclareMathAlphabet{\mathsfit}{\encodingdefault}{\sfdefault}{m}{sl}
\SetMathAlphabet{\mathsfit}{bold}{\encodingdefault}{\sfdefault}{bx}{n}

\title{Cascade: Token-Sharded Private LLM Inference}

\author{%
Rahul Thomas$^{1,2}$ \quad
Louai Zahran$^{1}$ \quad
Erica Choi$^{1,3}$ \quad
Akilesh Potti$^{1}$ \quad
Micah Goldblum$^{1,3}$ \quad
Arka Pal$^{1}$\thanks{Project lead and corresponding author.} \\[1.5ex]
$^1$Ritual \quad $^2$Stanford University \quad $^3$Columbia University \\[0.5ex]
\texttt{\{rahulthomas, louai, erica, akilesh, micah, arka\}@ritual.net}
}

\iclrfinalcopy 
\begin{document}

\maketitle

\begin{abstract}
As LLMs continue to increase in parameter size, the computational resources required to run them are available to fewer parties. Therefore, third-party inference services -- where LLMs are hosted by third parties with significant computational resources -- are becoming increasingly popular. However, third party inference raises critical concerns about user data privacy. To mitigate these risks, privacy researchers have developed provably secure schemes for third-party inference, such as Secure Multi-Party Computation (SMPC). However, SMPC protocols have significant computational and communication overhead, and do not scale to large models. In this work, we propose a new multi-party inference protocol, \newmethod{}, that avoids these punitive costs by leveraging sharding in the sequence dimension to maintain privacy, trading off cryptographic privacy guarantees for increased performance and scalability. We demonstrate that \newmethod{} is resistant to a generalization of a recent attack that is highly effective against other statistical privacy schemes, and that it is further resistant to learning-based attacks. As \newmethod{} is orders of magnitude faster than existing schemes, our findings offer practical solutions for secure deployment of modern state-of-the-art LLMs.
\end{abstract}

\section{Introduction}
\label{sec:introduction}


Due to recent advances in their abilities, Large Language Models (LLMs) have been increasingly used for a wide range of tasks, including but not limited to code generation, language processing, and complex reasoning \citep{zhu2024multilingualmachinetranslationlarge, kasneci2023chatgptgoodeducation, thirunavukarasu2023llms_medicine}. However, modern LLMs often contain hundreds of billions of parameters and require significant hardware resources to deploy. In particular, recent \emph{open-weights} models achieve cutting-edge performance \citep{deepseekai2025deepseekr1incentivizingreasoningcapability, qwen2025qwen25technicalreport} but remain too costly for individuals or organizations to run on their own. Thus, the demand for third-party LLM inference providers has grown significantly.  This raises critical privacy risks when data confidentiality is imperative, such as in areas with strict data privacy laws (e.g. GDPR in Europe), or in domains where sensitive individual information is revealed (e.g. healthcare). 

As a result, privacy researchers have focused on creating performant protocols for LLM inference, which do not reveal underlying information about the input prompt to the third-party provider. Many of these schemes rely on an approach called Secure Multi-Party Computation (SMPC), which involves multiple parties jointly performing inference, in such a way that no party can reconstruct the input from the information it receives \citep{yao1982protocolssecurecomputations, goldreich1987howtoplay}. Recently, various SMPC schemes have been formulated for for LLMs \citep{huang2022cheetah, hao2022iron, pang2023bolt, akimoto2023privformer, dong2023pumasecureinferencellama7b, li2024nimbussecureefficienttwoparty}, which rely on protocols based on additive-secret-sharing. These protocols are efficient for the bulk of LLM operations, such as matrix multiplication, but remain bottlenecked at non-linearities. Many works attempt to bypass this limitation by using SMPC-friendly approximations for non-linearities, such as piecewise polynomials. However, such approximations can impact downstream performance, and still remain far more expensive than direct computation of non-linearities. 

Therefore, recent works have sought to mitigate non-linearity costs by using statistical obfuscation approaches rather than cryptographic primitives. These schemes are significantly faster than SMPC schemes, but they can only guarantee statistical security, not theoretical security. For example, recent schemes \citep{zheng2024permllmprivateinferencelarge, yuan2024securetransformerinferenceprotocol, luo2024centaurbridgingimpossibletrinity} leveraged the permutation-equivariance properties of transformers \citep{xu2024permutationequivariancetransformersapplications} to obtain permutation-based schemes for private inference. Here, hidden states are revealed as permuted plaintext to the party performing the inference. These works justify security by referring to the extremely large permutation space, and concluding that the reversal of these permuted states to the original user prompts is infeasible.

However, this argument has recently been shown to be flawed. In recent work \citet{thomas2025attack} devise a reconstruction attack, the \textbf{vocab-matching attack}, that they show is able to near-perfectly decode input tokens from LLM hidden states. The authors extend the attack to further decode input tokens from various permutations of hidden states, breaking the security of three statistical permutation-based schemes \citep{zheng2024permllmprivateinferencelarge, yuan2024securetransformerinferenceprotocol, luo2024centaurbridgingimpossibletrinity}. 

\begin{figure}[t!]
  \centering
  \includegraphics[width=\linewidth]{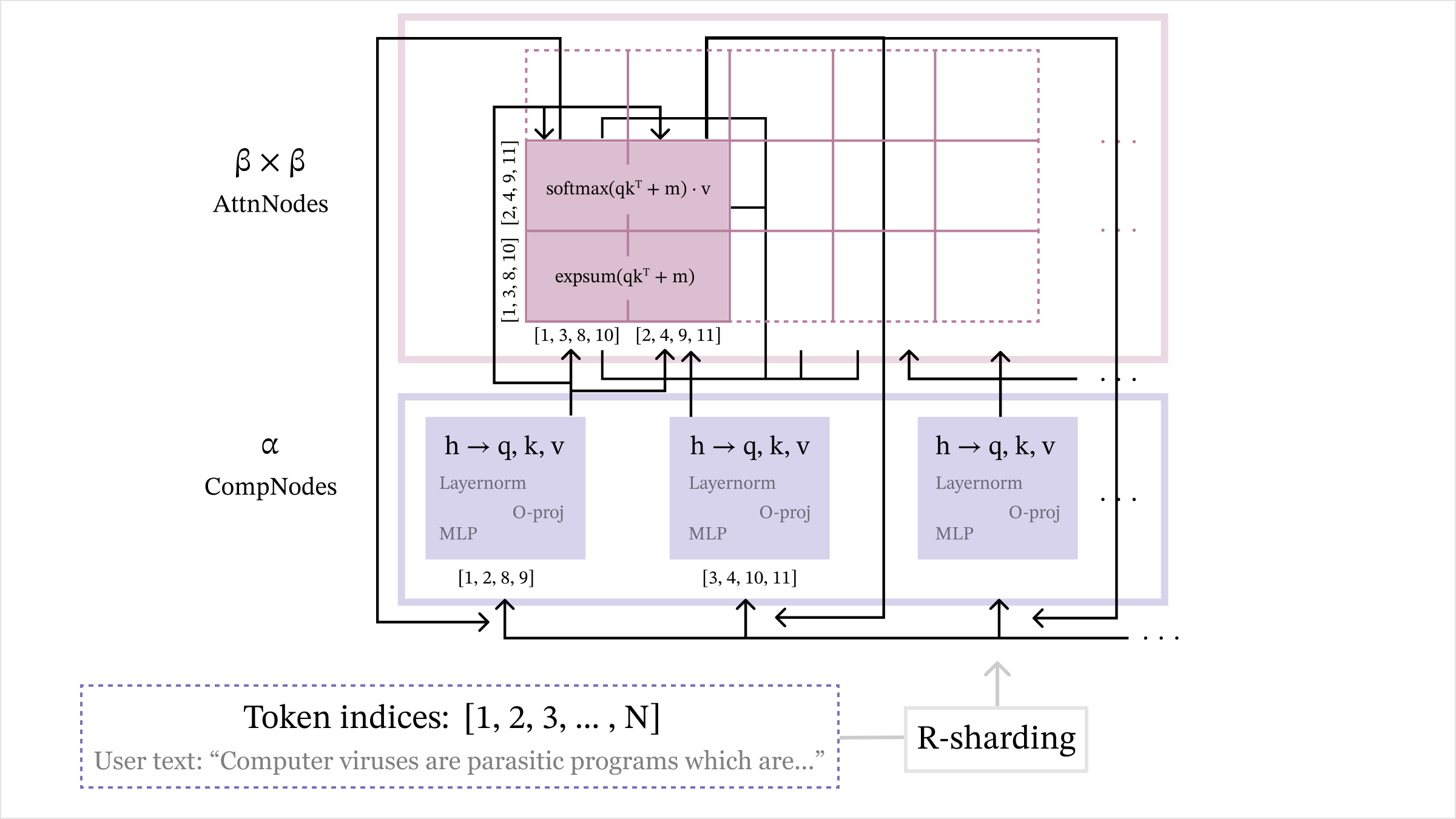}
  \caption{Schematic representation of our proposed privacy-preserving multi-party inference scheme, \textbf{\newmethod{}}.}
  \label{fig:figure_1}
\end{figure}

In this paper, we introduce a new multi-party scheme, \textbf{\newmethod{}}, that is resistant to the vocab-matching attack (see \cref{fig:figure_1}). Although this is a statistical scheme, unlike previous works, it leverages \emph{sharding} at the token level for obfuscation. While \newmethod{} does not have the rigorous privacy guarantees of cryptographic MPC schemes, it is much more efficient than them -- up to 100x faster overall, and with over 150x lower communication costs -- and presents a new paradigm in the trade-off between scalability and security.

The main contributions of our paper are:

\begin{enumerate}
  \item We generalize the vocab-matching attack introduced in \citet{thomas2025attack} to \textit{sharded} hidden state reversal. While the vanilla vocab-matching attack decodes full hidden states, our generalization is able to decode arbitrary slices of $N \times d$ hidden states along the token dimension $N$. 
  \item We introduce \newmethod{}, a multi-party statistical scheme for inference in the \textit{open-weights} setting that relies on sharding at the token level. We demonstrate that \newmethod{} is secure against our generalized vocab-matching attack for certain choices of sharding along the token dimension. We also show that \newmethod{} is resistant to existing learning-based reversal approaches in the literature \citep{wan2024informationleakageembeddinglarge, morris2023languagemodelinversion}.
  \item We demonstrate the efficiency and scalability of \newmethod{}. As our scheme does not employ cryptographic primitives like additive-secret-sharing or homomorphic encryption, it is significantly more lightweight than existing schemes, and is able to scale to modern state-of-the-art LLMs.
\end{enumerate}

\section{Setup \& Threat Model}

We consider the setting where a user $U$ wants to perform inference with an LLM model $M$ on some input prompt $\boldsymbol{x}$, which can be considered as an ordered sequence of tokens $[x_1, x_2, ..., x_N]$. We denote the size of the hidden state of the LLM by $d$, and the sequence length by $N$. 

As the user $U$ does not have the resources to perform the inference themselves, they rely on a set of third-parties $P_1, P_2, ..., P_K$. The model weights of $M$ are known to all parties. We consider the setting where each of the parties behaves \emph{semi-honestly}, a common assumption of past works \citep{zheng2024permllmprivateinferencelarge, luo2024centaurbridgingimpossibletrinity, dong2023pumasecureinferencellama7b, yuan2024securetransformerinferenceprotocol}. Semi-honest parties will follow the defined protocol faithfully, but may exploit any information that they receive during the execution of the protocol to attempt to recover the user's data.

Although it is not necessary for the setup and description of \newmethod{}, we further assume that the LLM model $M$ employs \textit{unidirectional} self-attention. This can be considered a worst-case assumption in security analysis, as the vocab-matching attack relies on this property to succeed.

\section{Related Work}
\label{sec:related_work}

Recent cryptographically-based SMPC schemes aim to alleviate the computational overhead of computing non-linearities in transformer architectures by devising new protocols for SMPC-unfriendly operations. In MPCFormer, \citet{li2023mpcformerfastperformantprivate} replace GeLU by a quadratic approximation and softmax with a normalized quadratic approximation. They employ knowledge distillation to fine-tune the model for downstream tasks, but they still see significant degradation in performance. To avoid the expensive operation of fine-tuning, Puma \citep{dong2023pumasecureinferencellama7b} formulates the first SMPC protocol for GeLU using piecewise polynomial fitting. However, this protocol still requires many multiplication and inequality protocol calls, and it is expensive. To improve the efficiency of GeLU, SecFormer \citep{luo2024secformer} replaces GeLU with a Fourier series approximation using segmented polynomials, and incorporates knowledge distillation as in MPCFormer. They are able to achieve similar runtimes as MPCFormer, with better downstream performance.

Most existing \textit{statistical} schemes that are related to our work are permutation-based. \citet{zheng2024permllmprivateinferencelarge} introduce the two-party scheme PermLLM, where hidden states are permuted at non-linear components to `safely` reveal elements to parties. \citet{yuan2024securetransformerinferenceprotocol} also develop STIP, a three-party scheme where the model server carries out inference on the user's input, without learning the proprietary model weights belonging to the model provider. This is accomplished with random permutations of both hidden states and model weights. In \citet{luo2024centaurbridgingimpossibletrinity}, the Cenatur scheme follows the same three-party model. However, they now avoid exposing the embedding lookup table and certain unpermuted intermediate results (e.g. attention logits), by incorporating additive-secret-sharing from SMPC at most stages of self-attention. In all of these schemes, permutations of hidden states are exposed to a party performing inference. \citet{thomas2025attack} use this observation to break the security of these schemes with their vocab-matching attack.


Later, to analyze the security of our statistical scheme, we will test reconstruction quality of learning-based attacks on exposed shards. The most relevant attacks in our setting are \citet{wan2024informationleakageembeddinglarge} and \citet{morris2023languagemodelinversion}. The former work focuses on reversal of hidden states, while the latter emphasizes logit distribution reversal. In both papers, the authors fine-tune a transformer-based architecture to reverse hidden states into their original input tokens. The latter paper does not assume any access by the adversary to model weights, while the former explicitly denotes the case of a model provider performing inference on user provided embeddings, and so is more analogous to our setting.

We will also test the security of our scheme against a generalization of the attack from \cite{thomas2025attack}, which is able to reverse input tokens from sharded hidden states. We describe this in greater detail in the next section.

\section{Generalized Vocab-Matching Attack}
\label{sec:walkthrough}

Here, we describe the vanilla attack from \cite{thomas2025attack}, where the adversary receives full layer $L$ hidden states $h_1,\ldots,h_N \in \mathbb{R}^d$ and attempts to recover the $N$-token input prompt. Then, we introduce a variant where the adversary only receives some of these $N$ hidden states but can perform a search over \textit{multiple} unknown tokens.

\subsection{Vanilla Attack}
\label{subsec:vanilla}

Suppose an LLM inference provider hosts a unidirectional LLM like Gemma-2-2B-IT and is given the following prompt by a user: "What is currently the most populated city in Spain?". After tokenization, this might be represented as the list [BOS, "What", "is", "currently", "the", "most", "populated", "city", "in", "Spain", "?", EOS], with BOS and EOS denoting the special tokens of begin-of-sequence and end-of-sequence respectively. These $12$ tokens have layer $0$ embeddings $e_1,\ldots,e_{12} \in \mathbb{R}^d$, and these give layer $L$ embeddings $h_1,\ldots,h_{12}\in \mathbb{R}^d$. Denoting the stacked decoder layers in Gemma-2-2B-IT as $\phi_1,\ldots,\phi_f$, which are all unidirectional, one sees for $\phi_{\leq L} = \phi_L \circ \ldots \circ \phi_2 \circ \phi_1$ that
$$\phi_{\leq L}(e_1)=[h_1] \in \mathbb{R}^{1 \times d}, \; \phi_{\leq L}(e_1,e_2)=[h_1,h_2] \in \mathbb{R}^{2 \times d}, \; \ldots, \; \phi_{\leq L}(e_1,\ldots,e_{12})=[h_1,\ldots,h_{12}] \in \mathbb{R}^{12 \times d}.$$
First, the provider takes the Gemma-2-2B-IT vocabulary $\mathcal{V}$ with size $V=256000$. For each token $v \in \mathcal{V}$ with embedding $e_v \in \mathbb{R}^d$, they compute $\phi_{\leq L}(e_v) \in \mathbb{R}^{1\times d}$, and then set $\widehat{e}_1$ as the $e_v$ which minimizes the L1 error $\|\phi_{\leq L}(e_v)_1-h_1\|_1$. Because $\phi_{\leq L}(e_1)=h_1$, then assuming there are no other collisions $\phi_{\leq L}(e_v)=h_1$, we will have $\widehat{e}_1=e_1$. Next, they compute $\phi_{\leq L}(\widehat{e}_1,e_v)\in \mathbb{R}^{2 \times d}$ for each $v \in \mathcal{V}$, and set $\widehat{e}_2$ as the $e_v$ which minimizes $\|\phi_{\leq L}(\widehat{e}_1,e_v)_2-h_2\|_1$. Again, without collisions, we have $\widehat{e}_2=e_2$. This continues until finally, they compute $\phi_{\leq L}(\widehat{e}_1,\ldots,\widehat{e}_{11},e_v) \in \mathbb{R}^{11 \times d}$, and set $\widehat{e}_{12}$ as the $e_v$ which minimizes $\|\phi_{\leq L}(\widehat{e}_1,\ldots,\widehat{e}_{11},e_v)_{12}-h_{12}\|_1$, which will be $e_{12}$. Now, the provider has true input embeddings $[\widehat{e}_1,\ldots,\widehat{e}_{12}]=[e_1,\ldots,e_{12}]$, which can be reversed into the $12$ input tokens via the lookup table. Here, non-collision is necessary for \textit{provable} success of the attack, as otherwise we may not have each $\widehat{e}_i=e_i$. 

\subsection{Sharding Generalization}
\label{subsec:generalized}

Now, we discuss a generalization of the vanilla attack above, with our psuedocode shown in \cref{alg:vocab_matching_attack_optimized_n}. This pertains to settings where the provider does not receive \textit{all} $N$ hidden states. For instance, suppose they only receive $h_3,h_5,h_{10}$. Can they still recover all input tokens?

The answer is negative. For instance, none of these $3$ hidden states are functions of $e_{11},e_{12}$, so one cannot always\footnote{One might be able to infer these from previous tokens for certain prompts. For instance, in the example prompt, "?" and EOS could be inferred from the rest of the prompt.} guarantee recovery of tokens $11$ and $12$. Nevertheless, our generalized attack, with sufficiently large computation cost, allows recovery of all tokens that the last hidden state $h_{10}$ depends on, meaning the first $10$ tokens. 

First, since the provider does not know $h_1$, they cannot carry out the first step of the vanilla attack to decipher $e_1$. However, if they generalize the first step, they can obtain something stronger from $h_3$: all of $e_1,e_2,e_3$. This is because $[h_1,h_2,h_3]=\phi_{\leq L}(e_1,e_2,e_3)$, so assuming non-collision\footnote{Actually, since we are matching only the last row to $h_3$, all that is needed is non-collision of the last row.} of the $V^3$ possible $3$-token forward passes $\{\phi_{\leq L}(e_{v_1},e_{v_2},e_{v_3})\}_{v_1,v_2,v_3\in \mathcal{V}}$, the provider can recover $e_1,e_2,e_3$ in at most $V^3$ forward passes. Next, from $[h_1,\ldots,h_5]=\phi_{\leq L}(e_1,\ldots,e_5)$ and non-collision of the $V^2$ possible $2$-token forward passes $\{\phi_{\leq L}(e_1,e_2,e_3,e_{v_4},e_{v_5})\}_{v_4,v_5\in \mathcal{V}}$, the provider can recover $e_4,e_5$ in at most $V^2$ forward passes. Finally, from $[h_1,\ldots,h_{10}]=\phi_{\leq L}(e_1,\ldots,e_{10})$ and non-collision of the $V^5$ possible $5$-token forward passes $\{\phi_{\leq L}(e_1,\ldots,e_5,e_{v_6},\ldots,e_{v_{10}})\}_{v_6,\ldots,v_{10}\in \mathcal{V}}$, the provider can recover $e_6,\ldots,e_{10}$ in at most $V^5$ forward passes. 

Now, the provider has recovered the first $10$ tokens in at most $V^3+V^2+V^5$ forward passes, where exponents $3,2,5$ correspond to gaps between indices of known hidden states. In general, if the provider gets $h_{i_1},\ldots,h_{i_k}$ with $1 \leq i_1 < \ldots < i_k \leq N$, then they can carry out this attack to reverse the first $i_k$ tokens in at most $V^{i_1} + V^{i_2-i_1} + \ldots + V^{i_k-i_{k-1}}$ forward passes. Although this attack appears powerful, the upper bound on forward passes can be quite large, so it is not always feasible. For instance, the dominating term in $V^3+V^2+V^5$ is $V^5 \approx 10^{27}$ for Gemma-2-2B-IT, which is entirely infeasible to search over for any adversary. In general, as $k \leq N =o(V)$ in practice, the cost bound $V^{i_1} + V^{i_2-i_1} + \ldots + V^{i_k-i_{k-1}}$ is dominated by the maximum-gap term $V^{g}$ with $g=\max_j (i_{j+1}-i_j)$. Thus, if $g$ is large enough so that the adversary \textit{cannot} perform $V^g$ forward passes, we would expect the attack to be infeasible. We define the minimum value of $g$ satisfying this as the \textbf{vocab-matching threshold $\rho$}.

\begin{definition}
    Let $t_\text{max}$ be the maximum value of $t$ for which an adversary has resources to perform $V^t$ forward passes. Then we define the vocab-matching threshold $\rho$ as $\rho := t_\text{max} + 1$.
\label{def:vma_threshold}
\end{definition}

Note that $\rho$ can be set on a case-by-case basis to suit the security demands of the user. Because $V \sim O(100000)$ in typical modern LLMs, the vocab-matching threshold of any adversary is likely no more than $\rho = 3$ in practice. 

Finally, we remark that the generalized attack requires stronger assumptions than the vanilla attack to succeed. For instance, in the vanilla attack, we only require non-collision to hold across $V$ forward passes at a time. In our example, however, we require it across up to $V^5$ forward passes, and in general, it is required across up to $V^g$ forward passes. However, as our generalization of the attack is considered primarily as a theoretical threat, we make the worst-case assumption that non-collision almost always holds in our security analysis.

\begin{algorithm}[t]
\caption{Generalized Vocabulary-Matching Attack}
\label{alg:vocab_matching_attack_optimized_n}
\begin{algorithmic}[1]
\INPUT Model $M$, indices $1 \leq i_1 < i_2 < \ldots <i_k \leq N$, layer $l$ hidden states $[h_{i_1},h_{i_2},\ldots,h_{i_k}]$, vocabulary $\mathcal{V}$
\OUTPUT Nearly decoded token sequence $\widehat{\boldsymbol{x}} = [\widehat{x}_1, \widehat{x}_2, \ldots, \widehat{x}_{i_k-1},\widehat{x}_{i_k}]$
\STATE Initialize empty sequence $\boldsymbol{\widehat{x}} \gets []$
\STATE $i_0 \gets 0$
\FOR{$j = 0$ to $k-1$}
    \STATE $\text{gap} \gets i_{j+1}-i_j$
    \STATE $\text{min\_dist} \gets \infty$
    \STATE $\text{best\_match} \gets \text{None}$
    \FOR{$v_{1},v_{2},\ldots,v_{\text{gap}} \in \mathcal{V}$}
        \STATE $g \gets M_{\leq l}([\widehat{\boldsymbol{x}},v_1,v_2,\ldots,v_{\text{gap}}])$ \COMMENT{Forward pass up to layer $l$} 
        \STATE $\text{dist} \gets \| g - h_{i_{j+1}}\|_1$ \COMMENT{Calculate L1 distance}
        \IF{$\text{dist} < \text{min\_dist}$}
            \STATE $\text{min\_dist} \gets \text{dist}$
            \STATE $\text{best\_match} \gets v$
        \ENDIF
    \ENDFOR
\ENDFOR
\STATE \textbf{return} $\widehat{\boldsymbol{x}}$
\end{algorithmic}
\end{algorithm}

\section{Cascade: Token-Sharded Multi-Party Inference}
\label{sec:cascade}

In \cref{sec:walkthrough}, we reviewed the vocab-matching attack from \cite{thomas2025attack} and showed how it could be extended to reverse tokens from token-sharded hidden states. However, we also saw that the cost of the generalized attack scaled exponentially with gaps between the revealed token indices, and so it is likely infeasible for sharded hidden states with sufficiently large token gaps. This motivates our defense to the generalized attack based on token-sharding of hidden states, which leads to a new statistical multi-party inference scheme: \textbf{\newmethod{}}.

Notably, \newmethod{} does not use cryptographic primitives; the computations are nearly unchanged from a standard forward pass, so almost no additional computational overhead is incurred. There is also no degradation of performance, as unlike SMPC schemes, no approximations need to be made to efficiently compute non-linearities. The scheme also does not require any trusted party interaction during inference. 

At a high level, \newmethod{} exploits the fact that only the self-attention mechanism in transformers has interaction among tokens in a sequence; for all other parts of the architecture, the tokens are treated like batch dimension elements. Therefore, the \emph{CompNodes} each receive only a subset of the tokens, and perform the bulk of batch operations in the LLM. Then, self-attention is performed by separate \emph{AttnNodes}, who each receive $Q,K,V$-projections of sharded hidden states from the CompNodes. The AttnNodes send partial attention outputs at each layer back to the CompNodes, who can reconstruct shards of the full attention output. 

In this section, we formally define the notation and protocols for \newmethod{}, \textit{without} initializing a specific sharding setup. In \cref{sec:security_analysis}, we will motivate the exact sharding scheme through a careful security analysis.

\subsection{Sharding and Notation}
\label{subsec:sharding_notation}

In multi-headed attention, we let $H$ and $H_{KV}$ be the attention head and key-value head (for grouped-query attention) counts, $N$ be the token count, $d_{emb}$ be the hidden dimension, and $d$ be the attention hidden dimension. There are three axes of sharding used along the token dimension: \textbf{(1)} the sharding of hidden states $\boldsymbol{h} \in \mathbb{R}^{N \times d_{emb}}$, \textbf{(2)} the sharding of query states $\boldsymbol{q} \in \mathbb{R}^{H \times N \times d}$, and \textbf{(3)} the sharding of key and value states $\boldsymbol{k},\boldsymbol{v} \in \mathbb{R}^{H_{KV} \times N \times d}$. These involve splitting token indices $[N]=\{1,2,\ldots,N\}$ into a union of disjoint subsets $\{R_i\}_{i=1}^\alpha$ for hiddens, $\{S_j\}_{j=1}^\beta$ for queries, and $\{T_k\}_{k=1}^\gamma$ for keys and values, where $\alpha,
\beta,\gamma$ are shard counts. We also shard the positional embeddings $\boldsymbol{p} \in \mathbb{R}^{N \times d_{emb}}$ and the attention mask $\boldsymbol{s} \in \mathbb{R}^{H \times N \times N}$, which are initialized pre-inference, as well as the masked logits $\boldsymbol{a} \in \mathbb{R}^{H \times N \times N}$ and attention output $\boldsymbol{o} \in \mathbb{R}^{N \times d_{emb}}$.

In practice, we can \textbf{symmetrize} $S$ and $T$ sharding. This arises from a pairwise coverage requirement: as $\{S_j\}_{j=1}^\beta$ and $\{T_k\}_{k=1}^\gamma$ cover $[N]$, each index pair $(x,y) \in [N] \times [N]$ lies in some $S_j \times T_k$. Because $S$ and $T$ sharding correspond to query and key/value sharding, this means for all such index pairs $(x,y)$, there is some AttnNode$_{jk}$ that holds both the $x$th query row and the $y$th key row. Observe that a node holding these rows has -- in the worst case -- the same information as if they hold the $y$th query and $x$th key rows. This is because the query and key matrices are linear projections of the same hidden states. Therefore, we can set $S$ and $T$ sharding equal at no loss of security. From here on, we replace $T$ with $S$ and $\gamma$ with $\beta$.

Our symmetrized notation for sharded intermediate LLM states at any layer is in \cref{tab:notation}. Note that the last six rows rely on tensors that are not directly shards of LLM states, but are derived from them, as in \cref{tab:derived_notation}. Here, max, softmax, and expsum are performed row-wise, with $\text{expsum}(\boldsymbol{x}):=\sum_i \text{exp}(x_i)$. Also, $\boldsymbol{e}^{S_k}$ only performs an expsum after subtracting row-wise maximums, as in the numerically stable subtract-max form of $\text{softmax}$.

\begin{table}[h]
\centering
\footnotesize
\setlength\tabcolsep{2pt}
\begin{tabular}{llll}
\toprule
\textbf{Tensor} & \textbf{Shape} & \textbf{Shards} & \textbf{Notation}\\
\midrule
$\boldsymbol{h},\boldsymbol{p},\boldsymbol{o}$ & $(N,d_{emb})$ & $(R_i,*)$ & $\boldsymbol{h}^{R_i},\boldsymbol{p}^{R_i},\boldsymbol{o}^{R_i}$ \\

$\boldsymbol{q},\boldsymbol{k},\boldsymbol{v}$ & $(H,N,d)$ & $(*,R_i,*)$ & $\boldsymbol{q}^{R_i},\boldsymbol{k}^{R_i},\boldsymbol{v}^{R_i}$ \\

$\boldsymbol{q},\boldsymbol{k},\boldsymbol{v}$ & $(H,N,d)$ & $(*,S_j,*)$ & $\boldsymbol{q}^{S_j},\boldsymbol{k}^{S_j},\boldsymbol{v}^{S_j}$ \\

$\boldsymbol{q},\boldsymbol{k},\boldsymbol{v}$ & $(H,N,d)$ & $(*,R_i \cap S_j,*)$ & $\boldsymbol{q}^{R_i\cap S_j},\boldsymbol{k}^{R_i \cap S_j},\boldsymbol{v}^{R_i\cap S_j}$ \\

$\boldsymbol{a},\boldsymbol{s}$ & $(H,N,N)$ & $(*,*,S_k)$ & $\boldsymbol{a}^{S_k},\boldsymbol{s}^{S_k}$ \\

$\boldsymbol{a},\boldsymbol{s}$ & $(H,N,N)$ & $(*,R_i,S_k)$ & $\boldsymbol{a}^{R_i S_k},\boldsymbol{s}^{R_iS_k}$ \\  

$\boldsymbol{a},\boldsymbol{s}$ & $(H,N,N)$ & $(*,S_j,S_k)$ & $\boldsymbol{a}^{S_j S_k},\boldsymbol{s}^{S_jS_k}$ \\  

$\boldsymbol{a},\boldsymbol{s}$ & $(H,N,N)$ & $(*,R_i \cap S_j,S_k)$ & $\boldsymbol{a}^{ (R_i \cap S_j) S_k},\boldsymbol{s}^{ (R_i \cap S_j) S_k}$ \\ 

$\boldsymbol{m}^{S_k},\boldsymbol{e}^{S_k}$ & $(H,N,1)$ & $(*,R_i,*)$ & $\boldsymbol{m}^{R_iS_k},\boldsymbol{e}^{R_iS_k}$ \\

$\boldsymbol{m}^{S_k},\boldsymbol{e}^{S_k}$ & $(H,N,1)$ & $(*,S_j,*)$ & $\boldsymbol{m}^{S_jS_k},\boldsymbol{e}^{S_jS_k}$ \\

$\boldsymbol{m}^{S_k},\boldsymbol{e}^{S_k}$ & $(H,N,1)$ & $(*,R_i \cap S_j,*)$ & $\boldsymbol{m}^{(R_i \cap S_j)S_k},\boldsymbol{e}^{(R_i \cap S_j)S_k}$ \\ 

$\boldsymbol{u}^{S_k}$ & $(H,N,|S_k|)$ & $(*,R_i,*)$ & $\boldsymbol{u}^{R_iS_k}$ \\

$\boldsymbol{u}^{S_k}$ & $(H,N,|S_k|)$ & $(*,S_j,*)$ & $\boldsymbol{u}^{S_jS_k}$ \\

$\boldsymbol{u}^{S_k}$ & $(H,N,|S_k|)$ & $(*,R_i \cap S_j,*)$ & $\boldsymbol{u}^{(R_i\cap S_j)S_k}$ \\

\bottomrule
\end{tabular}
\caption{Sharding for intermediate LLM states, where $*$ is a full slice. In slicing notation, $\boldsymbol{a}^{(R_i\cap S_j)S_k}=\boldsymbol{a}[:,R_i \cap S_j,S_k]$ and $\boldsymbol{m}^{(R_i\cap S_j)S_k}=\boldsymbol{m}^{S_k}[:,R_i \cap S_j,:]$, with $\boldsymbol{m}^{S_k}$ as in \cref{tab:derived_notation}.}
\label{tab:notation}
\end{table}

\begin{table}[h]
\centering
\footnotesize
\setlength\tabcolsep{2pt}
\begin{tabular}{lll}
\toprule
\textbf{Tensor} & \textbf{Shape} & \textbf{Formula}\\
\midrule
$\boldsymbol{m}^{S_k}$ & $(H,N,1)$ & $\max(\boldsymbol{a}^{S_k})$\\

$\boldsymbol{e}^{S_k}$ & $(H,N,1)$ & $\text{expsum} (\boldsymbol{a}^{S_k}- \boldsymbol{m}^{S_k})$\\

$\boldsymbol{u}^{S_k}$ & $(H,N,|S_k|)$ & $\text{softmax}(\boldsymbol{a}^{S_k})\boldsymbol{v}^{S_k}$\\
\bottomrule
\end{tabular}
\caption{Tensors derived from sharded states $\boldsymbol{a}^{S_k},\boldsymbol{v}^{S_k}$ in \cref{tab:notation}.}
\label{tab:derived_notation}
\end{table}

\subsection{Single Layer Pipeline}

We now describe how \newmethod{} operates for a single LLM layer. There are two types of nodes which hold our defined sharded states: \textbf{CompNodes} and \textbf{AttnNodes}. We initialize $\alpha$ CompNodes and $\beta^2$ AttnNodes, indexed as CompNode$_i$ and AttnNode$_{jk}$ for all $i \in [\alpha]$ and $j,k \in [\beta]$. \newmethod{} breaks single layer inference into the \textbf{pre-pass} by CompNodes, \textbf{attention-pass} by AttnNodes, and \textbf{post-pass} by CompNodes. We describe these in \cref{alg:comp_pre_pass}, \cref{alg:attn_pass}, and \cref{alg:comp_post_pass} in subsequent subsections. These form a single layer pass in \cref{alg:cascade_pass}.

\begin{algorithm}[ht]
\caption{Cascade Single Layer Forward Pass}
\label{alg:cascade_pass}
\begin{algorithmic}[1]
\INPUT $\boldsymbol{h}^{R_i}$ at layer $l$ from CompNode$_i$ for $1\leq i \leq \alpha$ 
\OUTPUT $\boldsymbol{h}^{R_i}$ at layer $l+1$ to CompNode$_i$ for $1\leq i \leq \alpha$ 
\FOR{$i=1$ to $\alpha$: CompNode$_i$}
    \STATE $\boldsymbol{q}^{R_i}, \boldsymbol{k}^{R_i}, \boldsymbol{v}^{R_i} \gets \textbf{pre\_pass}(\boldsymbol{h}^{R_i})$
    \let\OldAlgorithmicDo\algorithmicdo
    \renewcommand{\algorithmicdo}{AttnNode$_{jk}$ \textbf{gets}}
    \FOR{$j,k=1$ to $\beta$:}
        \STATE $\boldsymbol{q}^{R_iS_j},\boldsymbol{k}^{R_iS_k},\boldsymbol{v}^{R_iS_k}$
    \ENDFOR
    \let\algorithmicdo\OldAlgorithmicDo
\ENDFOR
\FOR{$j,k=1$ to $\beta$: AttnNode$_{jk}$}
    \STATE $\boldsymbol{m}^{S_jS_k}, \boldsymbol{e}^{S_jS_k},\boldsymbol{u}^{S_jS_k} \gets \textbf{attn\_pass}(\boldsymbol{q}^{S_j},\boldsymbol{k}^{S_k},\boldsymbol{v}^{S_k})$
    \let\OldAlgorithmicDo\algorithmicdo
    \renewcommand{\algorithmicdo}{CompNode$_{i}$ \textbf{gets}}
    \FOR{$i=1$ to $\alpha$:}
        \STATE $\boldsymbol{m}^{(R_i\cap S_j)S_k}, \boldsymbol{e}^{(R_i\cap S_j)S_k},\boldsymbol{u}^{(R_i\cap S_j)S_k}$
    \ENDFOR
    \let\algorithmicdo\OldAlgorithmicDo
\ENDFOR
\FOR{$i=1$ to $\alpha$: CompNode$_i$}
    \STATE $\boldsymbol{o}^{R_i} \gets \textbf{post\_pass}\{\boldsymbol{m}^{R_iS_k},\boldsymbol{e}^{R_iS_k},\boldsymbol{u}^{R_iS_k}\}_{k=1}^\beta$
    \STATE $\boldsymbol{h}^{R_i} \gets \textbf{mlp}(\boldsymbol{h}^{R_i} + \boldsymbol{o}^{R_i})$ \COMMENT{Residual and MLP}
\ENDFOR
\end{algorithmic}
\end{algorithm}

During \newmethod{} layer $l$ inference, each CompNode$_i$ begins with $\boldsymbol{h}^{R_i}$ at layer $l$, and ends with $\boldsymbol{h}^{R_i}$ at layer $l+1$. This output is the input to CompNode$_i$ for layer $l+1$ inference, so inference for all layers can follow \cref{alg:cascade_pass}.

After the last layer, CompNodes apply the LM head to get $R_i$-sharded logits, and the CompNode with the last token will use this to generate the next token's embedding\footnote{The logits can also be returned to the user if desired.}. This is sent back to the CompNode with the next token index in $R_i$. Then, \newmethod{} inference repeats to generate the next token. This can be sped up with KV-caching: see \cref{appendix:cost_optimization}.

Now, we describe the pre-pass, attention-pass, and post-pass components of \newmethod{}. All shard-specific slicing and concatenation operations are initialized in the node setup. Furthermore, we assume that all AttnNodes wait until all CompNodes finish the pre-pass to do the attention-pass, and all CompNodes wait until all AttnNodes finish the attention-pass to do the post-pass. 

\subsubsection{Pre-Pass}
At layer $l$, each CompNode$_i$ starts with the $R_i-$sharded hidden states $\boldsymbol{h}^{R_i}$, and applies layer normalization if necessary. Then, it $Q,K,V$-projects these to get the $R_i-$sharded query, key and value states $\boldsymbol{q}^{R_i},\boldsymbol{k}^{R_i},\boldsymbol{v}^{R_i}$. CompNode$_i$ then applies rotary or positional embedding to $\boldsymbol{q}^{R_i},\boldsymbol{k}^{R_i}$, using sharded positional embeddings $\boldsymbol{p}^{R_i}$ (the node can generate these upon setup to avoid any communication overhead, since it only depends on its index set $R_i$), and returns all of $\boldsymbol{q}^{R_i},\boldsymbol{k}^{R_i},\boldsymbol{v}^{R_i}$, as described in \cref{alg:comp_pre_pass}.

\begin{algorithm}[H]
\caption{CompNode$_i$ Single Layer Pre-Pass}
\label{alg:comp_pre_pass}
\begin{algorithmic}[1]
\INPUT $\boldsymbol{h}^{R_i},\boldsymbol{p}^{R_i}$
\OUTPUT $\boldsymbol{q}^{R_i},\boldsymbol{k}^{R_i}, \boldsymbol{v}^{R_i}$
\STATE $\boldsymbol{q}^{R_i} \gets \text{q\_proj}(\boldsymbol{h}^{R_i})$
\STATE $\boldsymbol{k}^{R_i} \gets \text{k\_proj}(\boldsymbol{h}^{R_i})$
\STATE $\boldsymbol{v}^{R_i} \gets \text{v\_proj}(\boldsymbol{h}^{R_i})$
\STATE $\boldsymbol{q}^{R_i} \gets \text{rotary\_pos\_emb}(\boldsymbol{q}^{R_i}, \boldsymbol{p}^{R_i})$
\STATE $\boldsymbol{k}^{R_i} \gets \text{rotary\_pos\_emb}(\boldsymbol{k}^{R_i}, \boldsymbol{p}^{R_i})$
\STATE \textbf{return} $\boldsymbol{q}^{R_i}, \boldsymbol{k}^{R_i}, \boldsymbol{v}^{R_i}$
\end{algorithmic}
\end{algorithm}

\subsubsection{Attention-Pass} After the pre-pass, each AttnNode$_{jk}$ receives shards $\boldsymbol{q}^{R_iS_j},\boldsymbol{k}^{R_iS_k},\boldsymbol{v}^{R_iS_k}$ from CompNode$_i$. By concatenating their rows over $1 \leq i \leq \alpha$, AttnNode$_{jk}$ obtains $\boldsymbol{q}^{S_j},\boldsymbol{k}^{S_k}, \boldsymbol{v}^{S_k}$. For example, for $\boldsymbol{q}^{S_j}$, concatenation is in the order in which one concatenates elements of \textit{sorted} $R_i \cap S_j$ over $1 \leq i \leq \alpha$ to get \textit{sorted} $S_j$. Then, AttnNode$_{jk}$ computes $\boldsymbol{a}^{S_jS_k} = \boldsymbol{q}^{S_j}(\boldsymbol{k}^{S_k})^T + \boldsymbol{s}^{S_jS_k}$, where matrix multiplication is per-head and we broadcast $H_{KV}$ to $H$. For the post-pass, AttnNode$_{jk}$ also stores row-wise maximums and subtract-max expsums $\boldsymbol{m}^{S_jS_k},\boldsymbol{e}^{S_jS_k}$. Finally, AttnNode$_{jk}$ takes the row-wise softmax and performs value multiplication to get $\boldsymbol{u}^{S_jS_k} = \text{softmax}(\boldsymbol{a}^{S_jS_k}) \boldsymbol{v}^{S_k}$. All of $\boldsymbol{m}^{S_jS_k},\boldsymbol{e}^{S_jS_k},\boldsymbol{u}^{S_jS_k}$ are returned, as in \cref{alg:attn_pass}.

\begin{algorithm}[H]
\caption{AttnNode$_{jk}$ Single Layer Attention-Pass}
\label{alg:attn_pass}
\begin{algorithmic}[1]
\INPUT $\boldsymbol{q}^{S_j},\boldsymbol{k}^{S_k},\boldsymbol{v}^{S_k},\boldsymbol{s}^{S_jS_k}$
\OUTPUT $\boldsymbol{m}^{S_jS_k},\boldsymbol{e}^{S_jS_k},\boldsymbol{u}^{S_jS_k}$
\STATE $\boldsymbol{k}^{S_k} \gets \text{repeat\_kv}(\boldsymbol{k}^{S_k})$
\STATE $\boldsymbol{v}^{S_k} \gets \text{repeat\_kv}(\boldsymbol{v}^{S_k})$
\STATE $\boldsymbol{a}^{S_jS_k} \gets \boldsymbol{q}^{S_j} (\boldsymbol{k}^{S_k})^T + \boldsymbol{s}^{S_jS_k}$
\STATE $\boldsymbol{m}^{S_jS_k} \gets \text{row\_max}(\boldsymbol{a}^{S_jS_k})$
\STATE $\boldsymbol{a}^{S_jS_k} \gets \text{exp}(\boldsymbol{a}^{S_jS_k} - \boldsymbol{m}^{S_jS_k})$
\STATE $\boldsymbol{e}^{S_jS_k} \gets  \text{row\_sum}(\boldsymbol{a}^{S_jS_k})$
\STATE $\boldsymbol{a}^{S_jS_k} \gets \boldsymbol{a}^{S_jS_k} / \boldsymbol{e}^{S_jS_k}$
\STATE $\boldsymbol{u}^{S_jS_k} \gets \boldsymbol{a}^{S_jS_k} \boldsymbol{v}^{S_k}$
\STATE \textbf{return} $\boldsymbol{m}^{S_jS_k},\boldsymbol{e}^{S_jS_k}, \boldsymbol{u}^{S_jS_k}$
\end{algorithmic}
\end{algorithm}

\subsubsection{Post-Pass}

Finally, after the attention-pass, each CompNode$_i$ receives $\boldsymbol{m}^{(R_i \cap S_j)S_k},\boldsymbol{e}^{(R_i \cap S_j)S_k},\boldsymbol{u}^{(R_i \cap S_j)S_k}$ from each AttnNode$_{jk}$, which are slices of its attention-pass outputs $\boldsymbol{m}^{S_jS_k},\boldsymbol{e}^{S_jS_k},\boldsymbol{u}^{S_jS_k}$ along the second-to-last dimensions. Then, for each fixed $1 \leq k \leq \beta$, CompNode$_i$ concatenates the rows of $\boldsymbol{m}^{(R_i \cap S_j)S_k},\boldsymbol{e}^{(R_i \cap S_j)S_k},\boldsymbol{u}^{(R_i \cap S_j)S_k}$ over all $1 \leq j \leq \beta$ to obtain $\boldsymbol{m}^{R_iS_k},\boldsymbol{e}^{R_iS_k},\boldsymbol{u}^{R_iS_k}$. Next, CompNode$_i$ aims to combine these results $\boldsymbol{m}^{R_iS_k},\boldsymbol{e}^{R_iS_k},\boldsymbol{u}^{R_iS_k}$ over $1 \leq k \leq \beta$ into the $R_i-$sharded (pre $O$-proj) output of attention. Using slicing notation, treating matrix multiplication as per-head, and broadcasting $H_{KV}$ to $H$, this is
$$\text{softmax}(\boldsymbol{a})[:,R_i,:] \boldsymbol{v} = \sum_{k=1}^\beta \text{softmax}(\boldsymbol{a})[:,R_i,S_k] \boldsymbol{v}[:,:,S_k] \in \mathbb{R}^{H \times |R_i| \times d}$$ by blocked matrix multiplication. The terms in the summation are not known to CompNode$_i$, since slicing here is performed post-softmax. To correct for this, observe that for a row vector $\boldsymbol{x} \in \mathbb{R}^N$ and any $1 \leq k \leq \beta$, again with slicing notation,
\begin{equation*}\text{softmax}(\boldsymbol{x} )[S_k] = \frac{\text{expsum}(\boldsymbol{x} [S_k])}{\sum_{l} \text{expsum}(\boldsymbol{x} [S_{l}])} \cdot \text{softmax}(\boldsymbol{x} [S_k]) \in \mathbb{R}^{|S_k|}.\end{equation*}
Thus, the above summation can be simplified as follows, with $\odot$ denoting elementwise multiplication:
\begin{align*}
\text{softmax}(\boldsymbol{a})[:,R_i,:] \boldsymbol{v} 
&= \sum_k \text{softmax}(\boldsymbol{a})[:,R_i,S_k] \boldsymbol{v}[:,:,S_k] \\
&= \frac{\sum_k \text{expsum}(\boldsymbol{a}[:,R_i,S_k]) \odot (\text{softmax}(\boldsymbol{a}[:,R_i,S_k]) \boldsymbol{v}[:,S_k])}{\sum_k \text{expsum}(\boldsymbol{a}[:,R_i,S_k])} \\
&=\frac{\sum_k \text{expsum}(\boldsymbol{a}^{R_iS_k}) \odot (\text{softmax}(\boldsymbol{a}^{R_iS_k}) \boldsymbol{v}^{S_k})}{\sum_k \text{expsum}(\boldsymbol{a}^{R_iS_k})} \\
&=\frac{\sum_k \text{exp}(\boldsymbol{m}^{R_iS_k}) \odot \text{expsum}(\boldsymbol{a}^{R_iS_k} - \boldsymbol{m}^{R_iS_k}) \odot (\text{softmax}(\boldsymbol{a}^{R_iS_k}) \boldsymbol{v}^{S_k})}{\sum_k \text{exp}(\boldsymbol{m}^{R_iS_k}) \odot \text{expsum}(\boldsymbol{a}^{R_iS_k} - \boldsymbol{m}^{R_iS_k})} \\
&= \frac{\sum_k \text{exp}(\boldsymbol{m}^{R_iS_k}) \odot \boldsymbol{e}^{R_iS_k} \odot \boldsymbol{u}^{R_iS_k}}{\sum_k \text{exp}(\boldsymbol{m}^{R_iS_k}) \odot \boldsymbol{e}^{R_iS_k}} \\
&= \frac{\sum_k \text{exp}(\boldsymbol{m}^{R_iS_k}-\boldsymbol{n}^{R_iS_k}) \odot \boldsymbol{e}^{R_iS_k} \odot \boldsymbol{u}^{R_iS_k}}{\sum_k \text{exp}(\boldsymbol{m}^{R_iS_k}-\boldsymbol{n}^{R_iS_k}) \odot \boldsymbol{e}^{R_iS_k}}
\end{align*}
with $\boldsymbol{n}^{R_i} \in \mathbb{R}^{H \times |R_i| \times 1}$ being the elementwise maximum of $\boldsymbol{m}^{R_iS_k} \in \mathbb{R}^{H \times |R_i| \times 1}$ over all $1 \leq k \leq \beta$. The fraction is elementwise divison, and $\text{expsum}$ is performed row-wise along the last dimension. This expression is numerically stable because each $\boldsymbol{m}^{R_iS_k} - \boldsymbol{n}^{R_iS_k}\leq 0$ and each $\boldsymbol{e}^{R_iS_k} \leq 1$. Now, each aggregate term in the numerator and denominator summations is known to the CompNode. In essence, the CompNode is performing a weighted average of concatenated AttnNode $\boldsymbol{u}$ results, with the weights also coming from AttnNodes $\boldsymbol{m},\boldsymbol{e}$ results. To get the final output of attention corresponding to row indices in $R_i$, the CompNode finally performs $O$-projection. \cref{alg:comp_post_pass} implements this.

\begin{algorithm}[H]
\caption{CompNode$_i$ Single Layer Post-Pass}
\label{alg:comp_post_pass}
\begin{algorithmic}[1]
\INPUT ${\boldsymbol{m}}^{R_iS_k}, {\boldsymbol{e}}^{R_iS_k}, {\boldsymbol{u}}^{R_iS_k}$ for $1 \leq k \leq \beta$
\OUTPUT $\boldsymbol{o}^{R_i}$
\STATE Initialize $\boldsymbol{o}^{R_i}$ with zeroes and shape like $\boldsymbol{u}^{R_iS_1}$
\STATE Initialize $\boldsymbol{w}^{R_i}$ with zeroes and shape like $\boldsymbol{e}^{R_iS_1}$
\STATE $\boldsymbol{n}^{R_i} \gets \text{elementwise\_max} \{\boldsymbol{m}^{R_iS_k}\}_{k=1}^\beta$
\FOR{$k=1$ to $\beta$}
    \STATE $\boldsymbol{w}^{R_i}  \gets \boldsymbol{w}^{R_i} + \text{exp}(\boldsymbol{m}^{R_iS_k}-\boldsymbol{n}^{R_iS_k}) \odot \boldsymbol{e}^{R_iS_k}$
    \STATE $\boldsymbol{o}^{R_i}  \gets \boldsymbol{o}^{R_i} + \text{exp}(\boldsymbol{m}^{R_iS_k}-\boldsymbol{n}^{R_iS_k}) \odot \boldsymbol{e}^{R_iS_k} \odot \boldsymbol{u}^{R_iS_k}$
\ENDFOR
\STATE $\boldsymbol{o}^{R_i} \gets \boldsymbol{o}^{R_i} / \boldsymbol{w}^{R_i}$
\STATE $\boldsymbol{o}^{R_i} \gets \text{o\_proj}(\boldsymbol{o}^{R_i})$
\STATE \textbf{return} $\boldsymbol{o}^{R_i}$
\end{algorithmic}
\end{algorithm}

\section{Security Analysis}
\label{sec:security_analysis}

We now examine the security properties of \newmethod{}. \newmethod{} does not employ cryptographic techniques, so we can only elucidate on statistical security. Nevertheless, we examine a wide range of security considerations below. The security of \newmethod{} is a function of its implementation parameters: the \emph{number of nodes} participating, as well as the \emph{sharding strategy} used, i.e. $\{R_i\}_{i=1}^\alpha$ and $\{S_j\}_{j=1}^\beta$.

To analyze information leakage, we examine isolated computational stages of nodes. In all stages (pre-pass, attention-pass, or post-pass) of \newmethod{}, no information is received \textit{during} the computation. Therefore, all input leakage comes from the stage initialization. Thus, any leakage must occur from the following sharded tensors:
\begin{align*}
    \text{CompNode}_i &\to \boldsymbol{h}^{R_i}, \{\boldsymbol{m}^{R_iS_k},\boldsymbol{e}^{R_iS_k},\boldsymbol{u}^{R_iS_k}\}_{k=1}^\beta, \\
    \text{AttnNode}_{jk} &\to \boldsymbol{q}^{S_j},\boldsymbol{k}^{S_k},\boldsymbol{v}^{S_k}.
\end{align*}

When more nodes participate, each has access to fewer tokens and hidden states, which makes reversing the full input more difficult. Even holding the node count fixed, the choice of sharding may also affect the difficulty of reversal. For example, is it more secure for a node to have access to $[h_1, h_5, h_9, h_{13}]$ or $[h_1, h_2, h_{12}, h_{13}]$?

As the space of sharding strategies is vast, we focus on \textbf{$\boldsymbol{(c,\delta)}$-sharding strategies}, where each sharded index set takes the form of a `clustered-arithmetic' or $(c,\delta)$-sequence. 

\begin{definition}
We say a subset of indices $[N]$ is a $(c,\delta)$-sequence if it takes the following form for some $i$:
\begin{align*}\{&i, i+1, \ldots, i+c - 1, \delta + i, \delta+i+1, \ldots, \delta+i+c-1,2\delta + i, 2\delta+i+1, \ldots, 2\delta+i+c-1,\ldots\}.
\end{align*}
\end{definition}

We focus on these strategies as they fulfill several security desiderata. We emphasize however that other strategies may be preferable to $(c,\delta)$-sharding depending on the use-case. Note that under $(c,\delta)$-sharding, the minimum number of CompNodes $\alpha$ needed to ensure all indices in $[N]$ are held by some node is given by $\alpha = \lceil \delta/c\rceil$.

We will examine leakage to CompNodes from $\boldsymbol{h}^{R_i}$ in \cref{subsubsec:security_layer_0} and \cref{subsubsec:security_layer_>_0}. We will analyze leakage from \textit{all} shards to CompNodes in \cref{subsubsec:security_other}. Finally, we examine leakage from $\boldsymbol{q}^{S_j},\boldsymbol{k}^{S_k},\boldsymbol{v}^{S_k}$ to AttnNodes in \cref{subsec:security_attn}.

\subsection{Layer 0 CompNode Hidden Security}
\label{subsubsec:security_layer_0}

We first consider the security of CompNodes at layer 0 of the scheme, who each hold some token embeddings of the input prompt $\boldsymbol{x}$. As embeddings are immediately reversible to their tokens through the lookup table, we conclude that Cascade should not be used when the security of every token is paramount. If individual token security is imperative, \newmethod{} can be integrated with SMPC as in \cref{appendix:cascade_smpc}.

\begin{figure}[h!]
    \centering
    \includegraphics[height=\textheight/4]{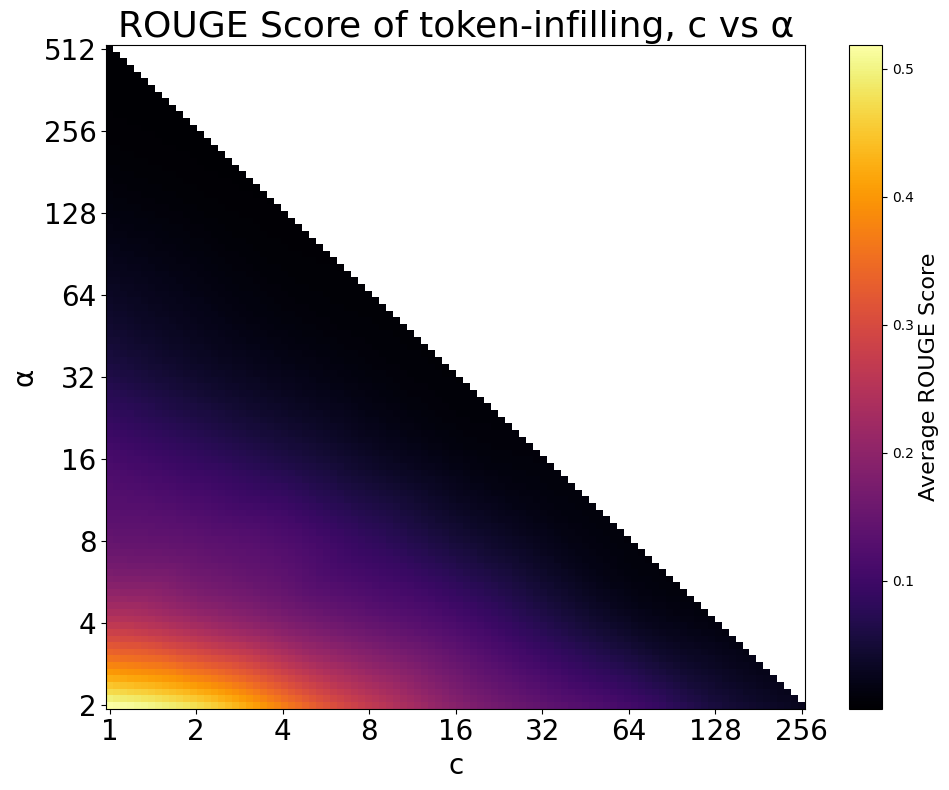}
    \caption{ROUGE-L scores for Layer $0$ token prediction using ModernBERT-Large, as a function of $c$, the number of `clusters' in the sharding scheme, and $\alpha$, the number of CompNodes. Higher $\alpha$ and higher $c$ tend towards lower ROUGE-L, increasing security.}
    \label{fig:c_alpha_rouge}
\end{figure}

Given that a CompNode has access to some of the $N$ token embeddings, say $[e_{n_1}, e_{n_2},\ldots, e_{n_t}]$, the possibility of reconstruction of the full prompt $x$ is theoretically lower bounded by the entropy of the distribution 
\begin{align*}
p(\{x_j : j \in [N] \setminus \{n_1,\ldots,n_t\}\} \mid x_{n_1}, x_{n_2}, \dots, x_{n_t})
\end{align*}
The true distribution cannot easily be computed, but we may approximate it using masked token infilling, as in the training of models like BERT \citep{devlin2019bertpretrainingdeepbidirectional}, RoBERTa \citep{liu2019robertarobustlyoptimizedbert}, and ModernBERT \citep{warner2024modernbert}. We use the recently released state-of-the-art ModernBERT-large to probe properties of this distribution under $(c,\delta)$-sharding. We use 200 samples from FineWeb-Edu to compute the ROUGE-L \citep{lin-2004-rouge} score over argmax-token generation. Our results are shown in \cref{fig:c_alpha_rouge}. We see that ROUGE-L score diminishes as both $c$ and $\alpha$ increase; good security seems to be achieved for $c,\alpha \geq 8$ and thus $\delta \geq 64$.


\subsection{Layer $>0$ CompNode Hidden Security}
\label{subsubsec:security_layer_>_0}

We now turn our attention to CompNode security of $\boldsymbol{h}^{R_i}$ at deeper layers of the LLM. Can we select a sharding strategy that defends against our attack outlined in \cref{sec:walkthrough}? Also, do we remain secure to learning-based attacks as in prior works \citet{wan2024informationleakageembeddinglarge} and \citet{ morris2023languagemodelinversion}?

\textbf{Vocab-Matching Attack}\hspace{1em} Our defense against vocab-matching is to ensure large token gaps in nodes. For a $(c,\delta)$-sharding scheme, in each shard, the distance from one `cluster' of indices to the next is $\delta - c + 1$. Therefore, when one carries out the generalized vocab-matching attack, the cost scales exponentially with the maximum gap, so it is on the order of $V^{\delta - c + 1}$. Note $\delta-c+1=(\alpha-1)c+1$, so regardless of the value of the vocab-matching threshold $\rho$, we can increase $\alpha$ or $c$ until $\delta-c+1\geq \rho$ to prevent vocab-matching. For $\rho = 3$, all $\alpha,c \geq 2$ satisfy this.

One objection to this line of argument is that if the adversary has a strong prior on the input text distribution, they can eliminate unlikely tokens from forward passes in the generalized attack, essentially only iterating over tokens in $\mathcal{V}_0 \subseteq \mathcal{V}$. However, the cost of this optimized attack is on the order of  $|\mathcal{V}_0|^{\delta-c+1}$, so it still scales exponentially. Thus, even if $|\mathcal{V}_0|$ is hundreds of times smaller than $|\mathcal{V}|\sim 100000$, large enough $\rho$ ensures that $\delta-c+1 \geq \rho$ remains secure.




\textbf{Learning-Based Attacks}\hspace{1em} We now consider learning-based reversal attacks. Although hidden states may not be as easily reversible as token embeddings, it may be the case that as attention propagates information among hidden states, the text-infilling task is easier at deeper layers of the LLM than at the purely textual level. We examine this hypothesis by fine-tuning Gemma-2-2B-IT and Llama-3.1-8B-Instruct on $(c,\delta)$-masked input sequences from FineWeb-Edu, with the target labels being the full input sequence. Note this is a `worst-case' scenario where the adversary knows the $c$ and $\delta$ parameters. We update both models to use a bidirectional mask, in line with the token-infilling nature of this task. We train until the evaluation loss on a held-out set converges over layer 1 representations from each model, and evaluate on layer 1 hidden states. Our approach is similar to  \citet{wan2024informationleakageembeddinglarge, morris2023languagemodelinversion}.

Our ROUGE-L scores for reconstructed Gemma-2-2B-IT prompts are shown in \cref{tab:c_alpha_rouge}. When $c, \alpha \geq 8$, we have a ROUGE-L score less than 0.25, indicating significant reconstruction difficulty. We also tested on $c=4, \alpha=16$ and $c=8, \alpha=24$ and obtained ROUGE-L scores of 0.173 and 0.144, supporting that security continues to improve by scaling $c$ and $\alpha$.

\begin{table}[h!]
\centering
\caption{ROUGE-L scores of text reconstruction from the hiddens of layer 1 of Gemma-2-2B-IT for various values of $c$ and $\alpha$ under $(c,\delta)$-sharding. Increasing $c$ or $\alpha$ results in worse reconstruction.}
\begin{tabular}{cccc}
\\ \toprule
 & \textbf{$\alpha=4$} & \textbf{$\alpha=8$} & \textbf{$\alpha=12$}\\
\midrule
$c=1$  &    0.701  &   0.467  &  0.349 \\
$c=4$  &    0.427  &   0.290  &  0.230 \\
$c=8$  &    0.355  &   0.222  &  0.191 \\
\bottomrule
\end{tabular}
\label{tab:c_alpha_rouge}
\end{table}

We similarly analyze BLEU scores \cite{papineni2002bleu} in \cref{tab:bleu-scores}. We find a similar trend as in \cref{tab:c_alpha_rouge}, observing that security improves as $c$ or $\alpha$ increases. Here, for $c\geq 4,\alpha=8$, we have a BLEU score between $0.1$ and $0.2$, indicating only marginal reconstructionability; and for $c\geq 4,\alpha=12$, the BLEU score lies below $0.1$, indicating nearly no reconstructionability.

\begin{table}[h!]
\centering
\caption{BLEU scores of text reconstruction from layer $1$ hiddens of Gemma-2-2B-IT for different values of $c,\alpha$ under $(c,\delta)$-sharding. Increasing $c$ or $\alpha$ results in worse reconstruction.}
\begin{tabular}{cccc}
\\ \toprule
 & \textbf{$\alpha=4$} & \textbf{$\alpha=8$} & \textbf{$\alpha=12$}\\
\midrule
$c=1$  &    0.537 & 0.229 & 0.130 \\
$c=4$  &    0.352 & 0.133 & 0.086 \\
$c=8$  &    0.246 & 0.123 & 0.083 \\
\bottomrule
\end{tabular}
\label{tab:bleu-scores}
\end{table}

We further examine the choice of hidden layer, and if Llama representations are decoded better, in \cref{appendix:reversal_layers_llama}. We find similar or better security across these other parameter choices.


\subsection{CompNode Full Security}
\label{subsubsec:security_other}

In \cref{subsubsec:security_layer_0} and \cref{subsubsec:security_layer_>_0}, we considered leakage to CompNodes from $\boldsymbol{h}^{R_i}$. Now, we consider how the shards $\boldsymbol{m}^{R_iS_k},\boldsymbol{e}^{R_iS_k},\boldsymbol{u}^{R_iS_k}$ impact security. Our main theorem shows that with this leakage, a scheme like $(c,\delta)$-sharding is, in some sense, \emph{necessary} to maintain security against vocab-matching: without sufficiently large gaps between consecutive clusters of indices, a variant of the attack can be carried out at Layer 0 to reveal additional tokens to a node. 

\begin{theorem}Suppose the attention mask $\boldsymbol{s}$ is unidirectional. Furthermore, denote the vocab-matching threshold as $\rho$, as defined in \cref{def:vma_threshold}. Then to prevent the generalized vocab-matching attack, for all $i \in [\alpha]$, each gap between clusters of consecutive indices in $R_i$ must have size $\geq \rho$.
\label{thm:compnode-full}
\end{theorem}

\begin{proof}Fix $i$, and denote $R_i=\{r_1,r_2,\ldots,r_m\}$ in ascending order. Consider any $k \in [\beta]$. Now, since $\boldsymbol{a}^{R_iS_k} = \boldsymbol{q}^{R_i}(\boldsymbol{k}^{S_k})^T + \boldsymbol{s}^{R_iS_k}$, and $\boldsymbol{s}$ is unidirectional, we see that the $l$th row of $\text{exp}(\boldsymbol{a}^{R_iS_k})\boldsymbol{v}^{S_k}$ is exactly $f(r_l,\{s \in S_k: s<r_l\})$, where we denote $$f(r,S):=\sum_{s \in S} \text{exp}\left(\boldsymbol{q}[:,r,:] (\boldsymbol{k}[:,s,:])^T\right) \boldsymbol{v}[:,s,:].$$
This $l$th row is known to CompNode$_i$ because we can write it in terms of known shards:
$$\text{exp}(\boldsymbol{a}^{R_iS_k})\boldsymbol{v}^{S_k} = (\text{softmax}(\boldsymbol{a}^{R_iS_k})\boldsymbol{v}^{S_k}) \odot \text{expsum}(\boldsymbol{a}^{R_iS_k}) = \boldsymbol{u}^{R_iS_k} \odot \boldsymbol{e}^{R_iS_k} \odot \text{exp}(\boldsymbol{m}^{R_iS_k}).$$
Thus, for each $r_l \in R_i$ and $k \in [\beta]$, we see that CompNode$_i$ knows $f(r_l,\{s \in S_k: s<r_l\})$. At Layer $0$, since $\boldsymbol{q},\boldsymbol{k},\boldsymbol{v}$ are linear projections of token embeddings, we see that $f(r,S)$ only depends on tokens with indices in the set $\{r\} \cup S$. Since CompNode$_i$ knows all tokens with indices $R_i$ from $\boldsymbol{h}^{R_i}$ at layer 0, then each $f(r_l,\{s \in S_k: s<r_l\})$ only depends on \textit{unknown} tokens at indices $\{s \in S_k: s<r_l\}\setminus R_i$. Thus, $f(r_{l+1},\{s \in S_k: s<r_{l+1}\})$ depends on the same unknown tokens as $f(r_l,\{s \in S_k: s<r_l\})$, and extra unknown tokens at $\{s \in S_k: r_l<s<r_{l+1}\}$, which we call the $(k,l)$-gap. 

If there are $<\rho$ tokens in a $(k,l)$-gap for some $l$, then following the generalized vocab-matching attack in \cref{sec:walkthrough}, CompNode$_i$ can perform a forward pass over all $V^{< \rho}$ candidate sequences of such unknown tokens in the $(k,l)$-gap and compute candidate values for $f(r_{l},\{s \in S_k: s<r_{l}\})$ and $f(r_{l+1},\{s \in S_k: s<r_{l+1}\})$, and then select the candidate sequence which allows these quantities to match their known values. This would allow them to recover all tokens in the $(k,l)$-gap. Thus, to prevent this vocab-matching attack, we need each $(k,l)$-gap to have size $\geq \rho$ or zero. This forces clusters of consecutive indices in $R_i$ to have size $\geq \rho$ gaps between them.
\end{proof}

Note that $(c,\delta)$-sharding satisfies the requirement of the theorem if and only if $\delta-c+1 \geq \rho+1$. Thus, this theorem induces a slightly stronger requirement than $\delta-c+1 \geq \rho$ from \cref{subsubsec:security_layer_>_0}, which only considered leakage from $\boldsymbol{h}^R_i$. In \cref{appendix:info_leakage}, we argue further that $(c,\delta)$-sharding with sufficiently large $\delta-c+1$ is \textit{sufficient} for \textit{Layer 0} security, by demonstrating a reduction to an intractable linear program.

\subsection{AttnNode Full Security}
\label{subsec:security_attn}

We now turn our attention to the analysis of AttnNode security. Recall that $S$ and $T$ sharding are set to be equal by the argument in \cref{subsec:sharding_notation}, and so in the following we discuss only $S$ sharding with the implication that $T$ receives the same treatment. 

Given a particular $R$ sharding, there are many possibilities for $S$ sharding.  One option is to set $S$ and $R$ sharding equal, and have each AttnNode$_{jk}$ receive the union of query shards from CompNode$_j$ and key/value shards from CompNode$_k$. However, this results in each AttnNode having access to twice as many $Q,K,V$-rows as each CompNode. To reduce such leakage to AttnNodes, we propose a further $m$-split of AttnNodes as follows.

To form $S$ sharding from $R$ sharding, we can let $\beta=m \alpha$, and partition each $R_i$ into $m$ shards $R_{i,1},\ldots,R_{i,m}$. Then the $S$ shards are $S_{m(i-1)+x} = R_{i,x}$ for all $1 \leq i \leq \alpha$ and $1 \leq x \leq m$. This ensures that pairwise coverage is maintained, but reduces the number of tokens (technically $Q,K,V$-projections of hidden state rows, but these are reversible to tokens in the worst case at layer $0$) that each AttnNode has access to by a factor of $m$. Using this split increases the value of $\beta^2$, the total number of AttnNodes, by a factor of $m^2$. 

There still remains a degree of freedom in deciding the exact choice of subdividing $R_i$ into the subsets $R_{i,x}$. Under the assumption that $R_i$ follows a $(c,\delta)$-sharding scheme, we propose that $R_{i,x}$ contains the elements of sorted $R_i$ at indices $x,x+\delta,\ldots,x+{(t-1)\delta}$, where $t = \frac{N}{c\alpha}$ and the split factor is $m=c$. 

For example, suppose that $\alpha=3$, $c=2$, $\delta=6$, and $N=18$. With a split of $m=2$, we have $\beta=m\alpha=6$. Then,
\begin{alignat*}{3}
    &R_1 = \{1,2,7,8,13,14\} \quad && R_2 = \{3,4,9,10,15,16\} \quad && R_3 = \{5,6,11,12,17,18\} \\
    &R_{1,1} = \{1,7,13\} \quad && R_{2,1} = \{3,9,15\} \quad && R_{3,1} = \{5,11,17\} \\
    &R_{1,2} = \{2,8,14\} \quad && R_{2,2} = \{4,10,16\} \quad && R_{3,2} = \{6,12,18\} \\
    &S_{1111} = \{1,7,13\} \quad && S_{1121} = \{1,3,7,9,13,15\} \quad && S_{1131} = \{1,5,7,11,13,17\} \\
    &S_{1112} = \{1,2,7,8,13,14\} \quad && S_{1122} = \{1,4,7,10,13,16\} \quad && S_{1132} = \{1,6,7,12,13,18\} \\
    &S_{2121} = \{3,9,15\} \quad && S_{2131} = \{3,5,9,11,15,17\} \quad && S_{2112} = \{2,3,8,9,14,15\} \\
    &S_{2122} = \{3,4,9,10,15,16\} \quad && S_{2132} = \{3,6,9,12,15,18\} \quad && S_{3131} = \{5,11,17\} \\
    &S_{3112} = \{2,5,8,11,14,17\} \quad && S_{3122} = \{4,5,10,11,16,17\} \quad && S_{3132} = \{5,6,11,12,17,18\} \\
    &S_{1212} = \{2,8,14\} \quad && S_{1222} = \{2,4,8,10,14,16\} \quad && S_{1232} = \{2,6,8,12,14,18\} \\
    &S_{2222} = \{4,10,16\} \quad && S_{2232} = \{4,6,10,12,16,18\} \quad && S_{3232} = \{6,12,18\}
\end{alignat*}

where we denote $S_{ixjy}=R_{i,x} \cup R_{j,y}$, which is the set of tokens that AttnNode$_{a,b}$ has access to for $a=m(i-1)+x,b=m(j-1)+y$. In other words, $R_i$ above are the sets of token indices that the CompNodes receive, and the $S_{ixjy}$ are the sets of token indices that the AttnNodes receive. Note that some $S_{ixjy}$ entries are not included above (there are $21$ listed, but $\beta^2=36$ AttnNodes) because they exactly match a listed entry by $S_{ixjy}= S_{jyix}$.

\textbf{Vocab-Matching Attack}\hspace{1em} Sharding $S$ in this way prevents vocab-matching, like in \cref{subsubsec:security_layer_>_0}. Indeed, as each $R_{i,x}$ has elements that are separated by $\delta$, and each $S_{ixjy}$ combines elements from two different $R_{i,x}$'s, then there cannot be $3$ consecutive elements in $S_{ixjy}$ if $\delta>2$. Also, the largest number of missing tokens between two elements of $S_{ixjy}$ (i.e. the largest `gap') is lower bounded by $\frac{\delta}{2}$. Therefore, for sufficiently large $\delta$, the vocab-matching attack is infeasible.

\textbf{Learning-Based Attacks}\hspace{1em} To test security against learning-based attacks, we conducted experiments with the above scheme for $m=\{2,3,4\}$, with $c=8$ and $\alpha=8$, and the same dataset and setup as in \cref{subsubsec:security_layer_>_0}. Due to computational constraints, we focus our experiments on Gemma-2-2B-IT on layer 1; we expect similar trends for Llama-3.1-8B-Instruct and other layers. We train a single model for all shard possibilities that arise from $S_{ixjy}$. Experiments are conducted with the same dataset and model setup as described . Our results are shown in \cref{tab:m_split_rouge}. We see that although $m=2$ results in a relatively higher ROUGE-L than that for the CompNodes in \cref{tab:c_alpha_rouge}, the score for $m=4$ is very similar; therefore, we recommend using $m \geq 4$ for security.

\begin{table}[h]
    \centering
    \caption{ROUGE-L scores for different values of splitting parameter $m$ on layer 1 of Gemma-2-2B-IT with $c=8, \alpha=8$. We see that the score for $m=4$ is similar to that of CompNodes in \cref{tab:c_alpha_rouge} for the same $c$ and $\alpha$.}
    \begin{tabular}{c c}
        \\
        \toprule
        $m$ & ROUGE-L \\
        \midrule
        2 & 0.3057 \\
        3 & 0.2643 \\
        4 & 0.2376 \\
        \bottomrule
    \end{tabular}
    \label{tab:m_split_rouge}
\end{table}

\subsection{Collusion Resistance}

Most SMPC schemes break when nodes share information. For example, in two-party additive secret sharing, a private value $x$ is split into shares $\langle x\rangle_0, \langle x\rangle_1$ held by parties $0,1$, and they can reconstruct $x = \langle x\rangle_0 + \langle x\rangle_1$ if they collude. In \cref{appendix:collusion}, we similarly analyze the collusion resistance properties of \newmethod{}. We find that colluding CompNodes and AttnNodes gain access to the \textit{union} of their token shards. Because unions of shards have smaller gaps between consecutive clusters, collusion directly impacts the feasibility of generalized vocab-matching, such as in \cref{thm:compnode-full}. Thus, although the exact collusion limit depends on security parameters like the vocab-matching threshold $\rho$ and the particular sharding scheme, \newmethod{} generally requires a large number of non-colluding nodes for security. 

\section{Communication and Computational Costs}

We now enumerate communication and computational costs associated with \newmethod{}. As in previous SMPC works \citep{li2023mpcformerfastperformantprivate, dong2023pumasecureinferencellama7b, li2024nimbussecureefficienttwoparty}, we assume perfect parallel transport in communication, a homogeneous node-wise bandwidth of $B$, and an inter-node latency of $\tau$. We denote $F$ as the number of bytes per element.

\subsection{Theoretical Costs}
\label{subsubsec:theory_costs}

We first provide theoretical estimates of costs in \newmethod{}.

\textbf{Computation}\hspace{1em} There is almost no floating point overhead in \newmethod{} relative to vanilla inference: see \cref{appendix:comp_costs}. We provide empirical evidence for this in \cref{subsubsec:benchmarking_runtime_comms}.

\textbf{Communication}\hspace{1em} We present single layer communication byte and time overhead formulae, derived in \cref{appendix:comm_costs}:
\begin{equation}\label{eq:comm_bytes}
\begin{aligned}
\text{CommBytes} &= \beta F (2dH +2dH_{KV} + 2H) \cdot N \\
\end{aligned}
\end{equation}
\begin{equation}\label{eq:comm_time}
\begin{aligned}
\text{CommTime}  &= 2\tau + \frac{\beta F d (H + 2H_{KV})}{B} \cdot \max_i |R_i| \\
                 &+ \frac{F (d+2) H}{B} \cdot \max_j |S_j|.
\end{aligned}
\end{equation}
Note the only sharding parameter \cref{eq:comm_bytes} depends on is $\beta$, and in fact it scales linearly with $\beta$. Thus, byte overhead is minimized when $\beta$ is minimized, which coincides with minimizing the AttnNode count. Furthermore, communication time is minimized when $\beta \max_i |R_i|$ and $\max_j |S_j|$ are minimal. For fixed $\alpha,\beta$, since $\{R_i\}_{i=1}^{\alpha}$ and $\{S_j\}_{j=1}^{\beta}$ partition $[N]$, then $\max_i |R_i| \geq \lceil N/\alpha \rceil$ and $\max_j |S_j| \geq \lceil N/\beta \rceil$. Here, equality holds when all $R_i$, and all $S_j$, are around the same size. Thus, $(c,\delta)$-sharding, which approximately divides tokens between CompNodes in equal quantity, has optimal communication time.


\subsection{Performance Experiments}
\label{subsubsec:benchmarking_runtime_comms}

We now evaluate the real-world performance of \newmethod{} through the distributed computing framework Ray \citep{moritz2018raydistributedframeworkemerging}. We run our experiments on Paperspace machines with 16 vCPU and 64GB RAM, with the CPU model being Intel Xeon Gold 6226R CPU @ 2.90GHz. All machines are colocated in the same region with an average bandwidth of 2 Gbps and latency of 0.38 ms. 

 We benchmark against two recent SMPC schemes for LLM inference, MPCFormer \citep{li2023mpcformerfastperformantprivate} and Puma \citep{dong2023pumasecureinferencellama7b}. For MPCFormer, we modify the Crypten implementation to use public rather than private weights, to match our open-weights setting. Puma data is taken from \citet{dong2023pumasecureinferencellama7b}, as it is built on SPU with its own set of optimizations. We perform inference (a single forward pass) on Bert-Base and Bert-Large with an input prompt of 128 tokens -- we repeat this measurement 100 times as there is run-to-run variability in the timing. Our results are shown in \cref{tab:bert_costs}.

We first compare \newmethod{} for $\alpha = 1$ without Ray against vanilla inference, to estimate protocol overhead. The mean runtime is 109ms vs. 91ms for vanilla inference for Bert-Base, and 320ms vs. 273ms for Bert-Large. Profiling shows this minor increase is due to the attention-score compilation step discussed in \cref{appendix:comp_costs}. Nevertheless, the mean runtime is within the 95\% confidence interval of vanilla inference in both cases.


Next, we compare the performance of \newmethod{} with $\alpha = 1$ and using Ray. As seen above, this is slower than not using Ray by around a factor of $3\times$. This slowdown can be attributed to framework-specific overhead, such as serialization. In other words, \newmethod{} is so efficient that the distributed-compute framework overhead now constitutes a significant proportion of its slowdown from vanilla inference, rather than protocol-specific overhead.


\begin{table}[ht]
    \centering
    \caption{Total runtime means and 95\% confidence intervals in seconds over 100 trials, for a single $128$-token prompt forward pass on Bert-Base and Bert-Large for MPCFormer, Puma, and various settings of \newmethod{}. Higher $\alpha$ corresponds to increased node counts and security.}
    \label{tab:bert_costs}
    \begin{tabular}{@{}lcc@{}}
        \toprule
        \textbf{Scheme} & \textbf{Bert-Base (s)} & \textbf{Bert-Large (s)} \\
        \midrule
        MPCFormer & 55.32 & 141.22 \\
        Puma      & 33.91 & 73.72 \\
        \toprule
        Cascade$_{\alpha=1,\text{no Ray}}$ & 0.11 [0.10, 0.13] & 0.32 [0.23,1.07]  \\
        Cascade$_{\alpha=1}$ & 0.32 [0.31, 0.36]  & 1.01 [0.97, 1.09] \\
        Cascade$_{\alpha=4}$ & 0.59 [0.51,0.69]  & 1.57 [1.44, 1.73] \\
        Cascade$_{\alpha=8}$ & 0.74 [0.62, 0.96] & 1.58 [1.27, 1.97] \\
        \toprule
        \textit{Vanilla} & 0.09 [0.08,0.12] & 0.27 [0.20, 0.99]  \\
        \bottomrule
    \end{tabular}
\end{table}

We further benchmark \newmethod{} with $\alpha = 4, 8$, using a $(c,\delta)$-sharding scheme with $c=1$ and no $m$-splits. We use clusters of $6$ and $18$ machines for $\alpha = 4$ and $8$. By testing the saturation of parallelization gains, we take an optimal $4$ cores per node. Performance in these cases is slower than for $\alpha = 1$, due to communication overhead. Still, \newmethod{} is $90\times$ faster than MPCFormer and $45\times$ faster than Puma for Bert-Large, even in its slowest setting.

Furthermore, although it is difficult to make strong conclusions from 3 data points, we observe that the increase in runtime for both models from $\alpha = 1$ to $\alpha = 8$ appears sublinear, even as the number of distinct machines increases superlinearly. This is expected due to parallel computation and communication across the nodes. Indeed, for Bert-Large, there is no statistically significant increase in runtime from $\alpha = 4$ to $\alpha = 8$, suggesting that it may even be saturating. In future work, we hope to run on even larger values of $\alpha$, with potential $m$-splitting, to confirm this hypothesis.

To support our theoretical estimates in \cref{subsubsec:theory_costs}, we also examined the total communication in the $\alpha = 1$ case using tshark. We found that the total communicated bytes are within $2\%$ of \cref{eq:comm_bytes}. The extra bytes can be attributed to Ray-specific metadata and a slight increase from the raw bytes size due to serialization. We present a comparative table of total communicated bytes for \newmethod{} versus MPCFormer and Puma in \cref{tab:bert_comms}. We see that even in the most expensive $\alpha = 8$ setting, \newmethod{} is $\sim160\times$ more efficient in total bytes transferred than MPCFormer, and $\sim140\times$ more efficient than Puma.

\begin{table}[ht]
    \centering
    \caption{Total gigabytes (GB) communicated for a single forward pass on Bert-Base and Bert-Large for MPCFormer, Puma, and \newmethod{} with various settings. A prompt length of 128 is used.}
    \label{tab:bert_comms}
    \begin{tabular}{@{}lcc@{}}
        \toprule
        \textbf{Scheme} & \textbf{Bert-Base (GB)} & \textbf{Bert-Large (GB)} \\
        \midrule
        MPCFormer & 12.089 & 32.577 \\
        Puma      & 10.773 & 27.246 \\
        Cascade$_{\alpha=1}$ & 0.009 & 0.025 \\
        Cascade$_{\alpha=4}$ & 0.038 & 0.101 \\
        Cascade$_{\alpha=8}$ & 0.076 & 0.203 \\
        \bottomrule
    \end{tabular}
\end{table}

\subsection{Scalability}

We also tested the scaling behaviour of \newmethod{} as a function of the model size. We again perform a forward pass with an input prompt of size 128 tokens, setting $\alpha=2$, and without $m$-splitting. We repeat this across 100 runs. Our results are shown in \cref{tab:cascade_scaling}. We observe that the runtime grows sublinearly with the model size, suggesting \newmethod{} will also scale well to the largest modern LLMs. We also note that Puma reported around $300$ seconds for a full forward pass on Llama-2-7B; \newmethod{} is over $20\times$ faster. 

\begin{table}[ht]
\centering
\caption{Total runtime means and 95\% confidence intervals in seconds over 100 trials, for a single $128$-token prompt forward pass on various sizes of LLM models with \newmethod{}, with $\alpha = 2$ and no m-splitting.  We observe sublinear increase in mean runtime as the model size grows, indicating that \newmethod{} is well suited to scale to the largest modern LLMs.}
\begin{tabular}{lccc}
\toprule
\textbf{Model Name} & \textbf{Model Size (Parameters)} & \textbf{Mean Runtime (s)} & \textbf{95\% Confidence Interval (s)} \\
\midrule
Bert-Base & 110M  & 0.70  & [0.62, 0.74] \\
Bert-Large & 335M  & 1.33  & [1.24, 1.46] \\
Llama-3.2-1B-Instruct & 1B  & 2.67  & [2.33, 2.96] \\
Llama-2-7B  & 7B  & 12.71 & [11.07, 14.07] \\
Llama-2-13B & 13B  & 22.72 & [20.58, 25.99] \\
\bottomrule
\end{tabular}

\label{tab:cascade_scaling}
\end{table}

\FloatBarrier

\section{Conclusion \& Future Work}

We generalized a recent attack for decoding LLM hidden states into their original user text in the setting of sharding in the token dimension. To defend against this, we introduced a novel multi-party scheme, \newmethod{}, that employs token sharding of hidden states. We showed that, under appropriate settings, \newmethod{} is resistant to our attack, as well as existing learning-based attacks in the literature. Finally, we benchmarked \newmethod{} on modern state-of-the-art LLMs, demonstrating significant improvements compared to existing multi-party inference schemes in terms of runtime, total communicated bytes, and scalability.

Future directions of work could examine the security and cost of other sharding strategies, particularly in unreliable network settings. This could include considerations of the asychronous setting, alongside other optimizations in \cref{appendix:cost_optimization}. Future work could also precisely optimize the security of Cascade for specific use cases, based on the form and frequency of private user data in prompts. We also encourage investigation into mixing Cascade with other SMPC schemes, such as our proposal in \cref{appendix:cascade_smpc}, to offer a flexible tradeoff between cryptographic security and efficiency. Such work could aim to prevent direct exposure of tokens to nodes, which is one of the main limitations of Cascade.

\bibliography{ref}
\bibliographystyle{plainnat}

\newpage

\appendix
\section{Integrating \newmethod{} with SMPC}
\label{appendix:cascade_smpc}

In \cref{subsubsec:security_layer_0}, we concluded that \newmethod{} should only be used when it is safe to reveal some number of tokens, but we mentioned that integration with SMPC could alleviate this requirement. Here, we describe how \newmethod{} can be fused with a number of SMPC schemes to improve token security at the expense of computational and communication cost. The idea is to form an \textbf{$L$-layered Cascade-SMPC split}, where a general SMPC scheme is executed on the first $L$ LLM layers and Cascade is carried out thereafter. 

We begin with the SMPC stage for layers $\leq L$. Denote the layer 0 input embeddings as $\boldsymbol{x}$, and the layer $L$ hidden states as $\boldsymbol{h}$. For the first $L$ layers, we execute any SMPC protocol $\Phi$ based on \textit{additive secret sharing}. Suppose $\Phi$ involves $t$ nodes $\mathcal{N}_1,\ldots,\mathcal{N}_t$. Then $\Phi$ begins by additively splitting $\boldsymbol{x} = \sum_{s=1}^t \boldsymbol{x}_s$, with each share $\boldsymbol{x}_s$ given to $\mathcal{N}_s$. By decomposing LLM layers into SMPC-friendly functions or approximations, $\Phi$ allows each node $\mathcal{N}_s$ to compute some additive share $\boldsymbol{h}_s$ of the layer $L$ hidden states, such that $\boldsymbol{h}=\sum_{s=1}^t \boldsymbol{h}_s$. A key ingredient of SMPC schemes based on additive secret sharing is \textit{computational indistinguishability}: no node gains any information about $\boldsymbol{x}$ during the execution of $\Phi$.

Once each $\mathcal{N}_s$ gets $\boldsymbol{h}_s$, we aim to execute Cascade for layers $> L$ in the LLM. We initialize a Cascade sharding scheme $\{R_i\},\{S_j\}$ and nodes $\{\text{CompNode}_i\},\{\text{AttnNode}_{jk}\}$, which are some \textit{superset} of the SMPC nodes\footnote{That is, unless there are more SMPC nodes than Cascade nodes. But this is usually never the case in practice, because most SMPC schemes based on additive secret sharing involve 2 or 3 parties.}. Then, for each $i$, each $\mathcal{N}_s$ sends the slice $\boldsymbol{h}_s^{R_i}=\boldsymbol{h}_s[R_i,:] \in \mathbb{R}^{|R_i| \times d}$ to CompNode$_i$. Finally, CompNode$_i$ adds its received sliced shares to get $\sum_{s=1}^t \boldsymbol{h}_s^{R_i} = \boldsymbol{h}^{R_i}$. Since each CompNode$_i$ has $R_i$-sharded layer $L$ hidden states, we can execute Cascade for all future layers of the LLM with the Cascade nodes.

In this process, by computational indistiguishability, SMPC nodes gain no information about the input, and thus have no direct token access. Likewise, computational indistiguishability ensures that sliced shares $\boldsymbol{h}_s^{R_i}$ give CompNode$_i$ no additional information about the input. That is, Cascade nodes no longer have \textit{direct} access to tokens: information leakage only comes from the usual exposed shards at layers $>L$. Thus, even though the execution of $\Phi$ on the first $L$ layers may be more expensive than Cascade, token security is improved. To determine the optimal choice of $L$, further analysis should be conducted on how information leakage from \textit{all} exposed Cascade shards varies across layers.

\section{CompNode Hidden Reversal Analysis on Layers \& Llama}
\label{appendix:reversal_layers_llama}

In \cref{subsubsec:security_layer_>_0}, we described experiments performed on the hidden states of Gemma-2-2B-IT, where a bidirectional-attention model was trained to reverse the sharded hidden states into the original text prompt. In \cref{tab:c_alpha_rouge}, we showed that the hiddens are largely secure to this attack for a suitable choice of $c$ and $\alpha$.

In this section, we first analyze if this is also true for Llama 3.1 8B-Instruct. We run a similar experimental setup as described in \cref{subsubsec:security_layer_>_0}, except we use Llama hidden representations, and we also use it as the reversal model; this also therefore tests if increasing the reversal model's capacity is a suitable method for improving sharded reconstruction. Due to the computational constraints of training with a larger model, we examine this only for $c=8, \alpha=8$ and $c=8, \alpha=12$. The reconstruction ROUGE-L scores are 0.1718 and 0.1443 respectively, significantly lower than those obtained with the same parameters for Gemma. We leave to future work the interesting question of whether this implies that Llama representations are inherently more resistant to decoding than Gemma representations.

Next, we analyze whether our results hold irrespective of the layer of the model used. We run additional experiments on the hiddens of layers 11 and 21 of Gemma-2-2B-IT. Our results are shown in \cref{tab:layer_rouge}. We see that there is no substantial difference in ROUGE-L score as the layer changes.

\begin{table}[h]
    \centering
    \caption{ROUGE-L scores of text reconstruction from the hiddens of various layers of Gemma-2-2B-IT for different $(c,\delta)$-sharding setups. We see that the reconstruction quality is similar across layers.}
    \begin{tabular}{c cc}
    \\
        \toprule
        \textbf{Layer} & $c=\alpha=4$ & $c=\alpha=8$ \\
        \midrule
        1  & 0.4268 & 0.2218 \\
        11  & 0.4627 & 0.2467 \\
        21 & 0.4021 & 0.2158 \\
        \bottomrule
    \end{tabular}
    \label{tab:layer_rouge}
\end{table}

\section{Computational Overhead Analysis}
\label{appendix:comp_costs}

In \cref{subsubsec:theory_costs}, we claimed that \newmethod{} has little overhead in computational costs compared to vanilla inference. We justify this statement below, by comparing CompNode and AttnNode steps against the vanilla forward pass.

Indeed, most operations performed by CompNodes will treat the (row) token dimension as the batch dimension. In the pre-pass, these are normalization and $Q,K,V$-projection; and in the post-pass, these are attention value compilation (most of \cref{alg:comp_post_pass}), $O$-projection, residual connection, and the MLP block. Except for attention value compilation, these steps all occur in the vanilla pass, so the CompNodes combined will perform the exact same operations as in vanilla inference: there are no extra computations performed.

The only extra operations thus come from \textbf{(a)} attention value compilation (linear weighting of partial attention outputs) by CompNodes in the post-pass, and \textbf{(b)} AttnNode floating point computations which do not appear in the vanilla pass, i.e. $\text{expsum}$s of shards of attention score rows. This is because all other steps of the \newmethod{} self-attention either treat the tokens as batch elements, or involve splitting up matrix multiplication into multiplication of sharded matrices; and the latter is blocked matrix multiplication, which does not inherently change the operations performed.

Now, \textbf{(a)} only involves $\sim H|R_i|d$ operations for each CompNode$_i$, since it involves a few steps of elementwise summation and multiplication of $H \times |R_i| \times d$ matrices (after broadcasting). Summing this over all $1 \leq i\leq \alpha$ gives $\sim \sum_i H|R_i|d = HNd$ extra operations. Also, \textbf{(b)} only involves $\sim H|S_j||S_k|$ operations for each AttnNode$_{jk}$ because $\text{expsum}$ is done over rows of an $H \times |S_j| \times |S_k|$ shard of attention scores. Summing over all $1 \leq  j,k \leq \beta $, we see this requires $\sim \sum_{j,k} H|S_j||S_k| = HN^2$ extra operations in total. This means the total AttnNode computation overhead is $\sim HN(d+N)$ operations. 

Importantly, this is cheaper than most computation-heavy steps in standard inference. Compared to the $\sim HN^2d$ operations from multiplication of $H \times N \times N$ attention scores with $H \times N \times d$ values, this overhead requires $\sim \frac{1}{N}+\frac{1}{d}$ times as many operations. Since $d$ is often in the hundreds, we can ensure for large $N$, say $N \geq 256$, that this ratio is quite small. Furthermore, if $N$ is not large, then the overhead is still limited compared to the $\sim HNd_{emb}d$ operations from $Q,K,V$-projection, since it requires $\sim \frac{1}{d_{emb}}+\frac{N}{d_{emb}}$ times as many operations and $d_{emb}$ is in the hundreds or thousands. Essentially, the choice of sharding does not significantly affect the total computational overhead, and this overhead is quite modest compared to the computations performed in a vanilla forward pass.

\section{Communication Analysis}
\label{appendix:comm_costs}

In \cref{subsubsec:theory_costs}, we gave the total communication byte and time overheads for performing a inference on a single layer of an LLM with \newmethod{}. Here, we provide a full justification of these equations. Like in \cref{appendix:comp_costs}, we assume symmetry of $S$ and $T$ sharding, so the superscript $T$ in sharded notation is replaced with $S$.

Recall that in each layer, there are two communication stages: \textbf{(A)} the CompNodes send sharded query, key, value matrices to the AttnNodes between pre-pass and attention-pass, and \textbf{(B)} the AttnNodes send sharded attention outputs and expsums to the CompNodes between attention-pass and post-pass. We operate under the assumption that all CompNodes synchronize before \textbf{(A)} and all AttnNodes synchronize before \textbf{(B)}, so that we can derive an exact expression for communication cost; this makes our communication cost derivation a worst-case analysis. See \cref{appendix:cost_optimization} for optimizations that can be made if this assumption is relaxed.

For single-layer inference, in stage \textbf{(A)}, CompNode$_i$ must send each of the $|R_i|$ rows of the $H \times |R_i| \times d$ query matrix $\boldsymbol{q}^{R_i}$ to some AttnNodes. In particular, for a row index $r \in R_i$, it sends the row $\boldsymbol{q}[:,r,:]$ of $\boldsymbol{q}^{R_i}$ to all AttnNodes$_{j_rk}$ with $1 \leq k \leq \beta$, where $j_r$ is the unique index satisfying $r \in S_{j_r}$. That is, CompNode$_i$ sends each of its $|R_i|$ rows to exactly $\beta$ AttnNodes. Since each row contains $Hd$ elements, then CompNode$_i$ must send out $\beta|R_i|Hd$ elements from sharded query states. A similar analysis shows CompNode$_i$ sends out $2\beta |R_i| H_{KV} d$ elements from sharded key and value states, so it sends out a total of $\beta|R_i|d (H+2H_{KV})$ elements. Summing this over all $i$ and noting $\sum_{i=1}^\alpha |R_i|=N$ gives the total bytes communicated in \textbf{(A)}:
$$\text{CommBytes}_A = \beta F d (H+2H_{KV}) \cdot N.$$ Assuming perfect parallel transport and uniform bandwidth $B$ across nodes, i.e. all communication overhead comes from CompNode with the most elements to send (plus latency $\tau$), the communication time in stage \textbf{(A)} is
$$\text{CommTime}_A = \tau + \frac{\beta F d (H + 2H_{KV})}{B} \cdot \max_i |R_i|.$$
Next, in stage \textbf{(B)}, each AttnNode$_{jk}$ must send each CompNode$_i$ some rows of its partial post-value attention outputs $\boldsymbol{u}^{S_jS_k}$, partial attention score row maximums $\boldsymbol{m}^{S_jS_k}$, and partial attention score row subtract-max expsums $\boldsymbol{e}^{S_jS_k}$. These matrices are of shapes $H \times |S_j| \times d,H \times |S_j| \times 1,H \times |S_j| \times 1$, respectively, and CompNode$_i$ receives $|R_i \cap S_j|$ out of the $|S_j|$ rows from each. This means the total number of elements that AttnNode$_{jk}$ sends to all CompNodes is 
$$(d+2)H \cdot \sum_{i=1}^\alpha |R_i \cap S_j| =(d+2) H \cdot |S_j|.$$
Since $\sum_{j,k=1}^\beta |S_j| = \beta \sum_{j=1}^\beta |S_j| = \beta N$, this means the total number of bytes sent by all $\beta^2$ AttnNodes is
$$\text{CommBytes}_B = \beta F (d+2) H \cdot N.$$
And, again under the parallel transport and uniform bandwidth assumption, the communication time in \textbf{(B)} is
$$\text{CommTime}_B = \tau + \frac{F (d+2) H}{B} \cdot \max_j |S_j|.$$
Combining these costs, we obtain the following total communication byte and time overheads for a single layer:
\begin{align*}
\text{CommBytes} &= \beta F (2dH +2dH_{KV} + 2H) \cdot N, \\
\text{CommTime}  &= 2\tau + \frac{\beta F d (H + 2H_{KV})}{B} \cdot \max_i |R_i| + \frac{F (d+2) H}{B} \cdot \max_j |S_j|.
\end{align*}

Finally, we compute the number of communication rounds per layer. Stage \textbf{(A)} has each of the $\alpha$ CompNodes send results to at most $\beta^2$ AttnNodes, which is at most $\alpha\beta^2$ rounds. Stage \textbf{(B)} has each of the $\beta^2$ AttnNodes send results to at most $\alpha$ CompNodes, which is at most $\alpha\beta^2$ rounds. In total, the rounds per layer are bounded above by $2\alpha\beta^2$. This can be quite large, but we can guarantee a tighter upper bound if our scheme involves $(c,\delta)$-sharding for CompNodes with $m$-splitting of AttnNodes. Here, $\beta=m \alpha$ since each of the $\alpha$ shards in $\{R_i\}_{i=1}^\alpha$ is split into $m$ pieces to form $\{S_j\}_{j=1}^\beta$. Each CompNode sends results to $m\beta$ AttnNodes, and each AttnNode sends results to $1$ CompNode, so there are $m\alpha\beta + \beta^2 = 2\beta^2$ rounds. Essentially, the number of rounds scales linearly with the number of AttnNodes.

\section{Sufficiency of $(c,\delta)$-Sharding}
\label{appendix:info_leakage}

Now, we provide an argument that only considering Layer 0 shards, when $R$ and $S$ sharding both follow a $(c,\delta)$-sharding scheme with sufficiently large $c,\delta$, it is intractable for any CompNode to recover the input.

To do this, we will demonstrate a reduction to a variant of the subset sum problem with vectors, which is NP-complete. Our reduction relies on two assumptions \textbf{(B1)}, \textbf{(B2)}, which we highlight shortly. We begin by fixing $i \in [\alpha]$, so that $R_i=\{ic,ic+1,\ldots,ic+c-1,ic+\delta,ic+\delta+1,\ldots,ic+\delta+c-1,\ldots\}$. As we mentioned earlier, CompNode$_i$ has access to $\boldsymbol{m}^{R_iS_1},\boldsymbol{e}^{R_iS_1},\boldsymbol{u}^{R_iS_1},\ldots,\boldsymbol{m}^{R_iS_\beta},\boldsymbol{e}^{R_iS_\beta},\boldsymbol{u}^{R_iS_\beta}$ and $\boldsymbol{h}^{R_i}$ at Layer 0. Knowledge of the latter is equivalent to knowledge of tokens at indices $R_i$, so we consider leakage from the former triples. In fact, at Layer 0, note each triple $\boldsymbol{m}^{R_iS_k},\boldsymbol{e}^{R_iS_k},\boldsymbol{u}^{R_iS_k}$ only depends on tokens with indices in $R_i \cup S_k$. Since all tokens in $R_i$ are known to CompNode$_i$, then this triple only depends on \textit{unknown} tokens $S_k \setminus R_i$. Across all $k$, these sets are disjoint. Thus, under the following assumption, we can conclude that the reversal problems from the $\beta$ different triples are \textit{independent}.

\textbf{(B1)} For any $k_1 \neq k_2$, CompNode$_i$ has independent priors on tokens with indices in $S_{k_1}$ and tokens with indices in $S_{k_2}$.

Thus, we now only need to consider leakage from \textit{one} triple $\boldsymbol{m}^{R_iS_k},\boldsymbol{e}^{R_iS_k},\boldsymbol{u}^{R_iS_k}$. Explicitly, $S_k=\{kc,kc+1,\ldots,kc+c-1,kc+\delta,kc+\delta+1,\ldots,kc+\delta+c-1,\ldots\}$. So, if $k=i$ and $S_k=R_i$, then all these shards are functions only of \textit{known} tokens at indices $R_i$, and no information is leaked to CompNode$_i$. Thus, without of loss of generality, we now assume $k<i$, as $k>i$ is similar. From now on, when we say a shard depends on tokens, we ignore known tokens.

Recall that $\text{exp}(\boldsymbol{a}^{R_iS_k})\boldsymbol{v}^{S_k}=\boldsymbol{u}^{R_iS_k} \odot \boldsymbol{e}^{R_iS_k} \odot \text{exp}(\boldsymbol{m}^{R_iS_k})$, and $\text{expsum}(\boldsymbol{a}^{R_iS_k})=\boldsymbol{e}^{R_iS_k} \odot \text{exp}(\boldsymbol{m}^{R_iS_k})$ by definition. Thus, this triple reveals the exact same information as the triple $\boldsymbol{m}^{R_iS_k},\text{expsum}(\boldsymbol{a}^{R_iS_k}),\text{exp}(\boldsymbol{a}^{R_iS_k})\boldsymbol{v}^{S_k}$, which have shapes $(H,|R_i|,1),(H,|R_i|,1),(H,|R_i|,d)$. We concatenate the last two to form $\boldsymbol{b}^{R_iS_k}$ of shape $(H,|R_i|,d+1)$, so our original triple is equivalent to $\boldsymbol{m}^{R_iS_k},\boldsymbol{b}^{R_iS_k}$. Now, note that because $S$ and $R$ are both $(c,\delta)$-sharding and the attention mask is unidirectional, the first $c$ out of $|R_i|$ rows of $\boldsymbol{a}^{R_iS_k}$ will have $c$ elements followed by all $-\infty$, with the nonzero $c$ only depending on tokens at indices $kc,\ldots,kc+c-1$, respectively. The next $c$ rows have $2c$ elements that depend on tokens at indices $kc,\ldots,kc+c-1,kc+\delta,kc+\delta+1,\ldots,kc+\delta+c-1$, followed by all $-\infty$. A similar pattern holds for each next $c$ rows. Thus, the $|R_i|$ rows of $\boldsymbol{b}^{R_iS_k}$ follow a similar token dependence pattern: the first $c$ depend on tokens $kc,\ldots,kc+c-1$, the next $c$ depend on tokens $kc,\ldots,kc+c-1,kc+\delta,kc+\delta+1,\ldots,kc+\delta+c-1$, and so on. 

Now, for each position $x \in [N]$, we define a list of $V$ possible concatenations $[k_x,v_x] \in \mathbb{R}^{H \times 2d}$, where $k_x$ denotes the $x$th row across all heads of the layer 0 query states, and likewise for $v_x$. Denote this list as $\mathcal{A}_x$, which is distinct for different values of $x$ due to the effects of positional embeddings; it is indexed in the same order as the vocabulary. For each $r \in R_i$ and position $x \in [N]$ with $x\leq r_i$, we denote $\mathcal{B}_{r,x}$ as the list of $(d+1)$-dimensional vectors whose first entry is $\text{exp}(\boldsymbol{q}[:,r,:] k_x^T)$ and last $d$ entries are $\text{exp}\left(\boldsymbol{q}[:,r,:] k_x^T\right) v_x$, over all $[k_x,v_x] \in \mathcal{A}_x$. Note this is also of size $V$, and is indexed in the same order as the vocabulary. It is simple for a node to compute $\mathcal{B}_{r,x}$ for all $r \in R_i$ and $x \leq r_i$, as they know $\boldsymbol{q}[:,r,:]$, and can directly compute and iterate through $\mathcal{A}_x$.

The key point, building on our observation from earlier, is that $(d+1)$-dimensional rows of $\boldsymbol{b}^{R_iS_k}$ are sums of vectors from sets $\mathcal{B}_{r,x}$. In particular, there are $i_1,\ldots,i_c$ such that for all $1 \leq r \leq c$, the $r$th row equals $\mathcal{B}_{r,kc}[i_1]+\ldots+\mathcal{B}_{r,kc+c-1}[i_c]$. Then, there are $i_{c+1},\ldots,i_{2c}$ such that for all $c+1 \leq r \leq 2c$, the $r$th row equals $\mathcal{B}_{r,kc}[i_1]+\ldots+\mathcal{B}_{r,kc+c-1}[i_c] + \mathcal{B}_{r,kc+\delta}[i_{c+1}]+\ldots+\mathcal{B}_{r,kc+\delta+c-1}[i_{2c}]$. A similar pattern holds for $2c+1 \leq r \leq 3c$, and so on. The final indices $i_1,i_2,\ldots$ correspond to the positions of input tokens in the vocabulary, for all tokens with indices in $S_k$. 

Therefore, reversal of the input from $\boldsymbol{b}^{R_iS_k}$ alone is at least as difficult as the following problem: given $c$ different sets, each containing $V$ vectors derived from token and position embeddings, select some vector from each set so that they sum to a given vector. This is a variant of the subset sum problem for vectors, which is computationally intractable for large enough $V,c$. Particularly, since $V$ is in the hundreds of thousands, this is likely already intractable for $c \geq 8$.

Finally, we consider additional leakage from $\boldsymbol{m}^{R_iS_k}$. In the worst case, this reveals at most one of the tokens with indices in $S_k$ for each $k$, as a maximum of a row of $\boldsymbol{a}^{R_iS_k}$ reveals at most one element of the row, with upper bound constraints placed on the other elements. Given the large number of possibilities for these elements, it is unlikely that the upper bound constraints provide much information, which we incorporate in the assumption below. Therefore, leakage from $\boldsymbol{m}^{R_iS_k}$ is at worst equivalent to effectively reducing the parameter $c$ above by one, since we have one less set to select a vector from. This means if $c \geq 8$ was secure for reversal from $\boldsymbol{b}^{R_iS_k}$, we need $c \geq 9$ to ensure security for reversal from $\boldsymbol{b}^{R_iS_k},\boldsymbol{m}^{R_iS_k}$. Note that the variant of the subset sum problem for reversal from $\boldsymbol{b}^{R_iS_k}$, together with the inequality constraints from $\boldsymbol{m}^{R_iS_k}$ leakage, form a linear program. So, we have implicitly used the following assumption to claim security.

\textbf{(B2)} The linear program described above is computationally intractable.

Therefore, assuming choices of $c,\delta$ which satisfy \textbf{(B1)} and \textbf{(B2)}, we have shown \newmethod{} is completely secure for any CompNode only executing Layer 0. In practice, these assumptions may not always hold. In fact, learning-based attacks attempt to violate the independence assumption of \textbf{(B1)}, through token-infilling between gaps of a $(c,\delta)$-sequence. Furthermore, even as the subset sum problem is NP-complete, this does not mean that the particular choice of sets in our setup are not susceptible to, say, an approximation algorithm. To truly test \textbf{(B2)}, future work should explicitly enumerate the linear program and test the efficacy of state-of-the-art solvers on it, perhaps with token-infilling priors integrated. Nevertheless, the above explicit reduction shows that for large enough $c,\delta$, it is quite likely that \textbf{(B1)} and \textbf{(B2)} nearly hold.

\section{Cost Optimizations}
\label{appendix:cost_optimization}

In \cref{sec:cascade}, we gave a high-level overview of \newmethod{}, and deferred discussions about optimization. Here, we discuss a few cost and communication optimizations.

\textbf{Caching}\hspace{1em} After a new token is generated in \newmethod{}, the CompNode holding that token will send it back to \textit{one} of the existing CompNodes, and single-token generation will repeat to get the next token. To speed up generation after the first new token, the CompNodes and AttnNodes can store their partial intermediate states, and only the $1$ CompNode and $\beta$ AttnNodes associated with the most recent token will need to participate in the single-token generation: this means KV-caching naturally extends to \newmethod{}. Formally, suppose $n$ is the index of the most recently generated token, and it belongs to the hidden shard $R_{i}$ and the query shard $S_j$. Only CompNode$_i$ needs to perform new computation in generating the $(n+1)$st token, along with each AttnNode$_{jk}$ for $1 \leq k \leq \beta$: this is because only these AttnNodes require the $n$th query row. Furthermore, these $\beta+1$ nodes, having stored intermediate results from previous forward passes, can avoid repeat computation of attention scores and earlier hidden states. Essentially, this results in token-sharded KV-caching.

\textbf{Symmetry Reduction}\hspace{1em} We see that AttnNode$_{jk}$ and AttnNode$_{kj}$ actually have the exact same information in the worst-case: they both have access to indices $S_j \cup S_k$. Thus, at no loss of security, we can combine AttnNode$_{jk}$ and AttnNode$_{kj}$ into one node, thereby approximately halving the number of AttnNodes required, and reducing communication byte overhead.

\textbf{Synchronization}\hspace{1em} A key assumption in our communication analysis from \cref{appendix:comm_costs} was that nodes synchronize between stages. That is, AttnNodes wait until they all finish before sending information to CompNodes in parallel; and likewise for the CompNodes sending information to AttnNodes. But in practice, depending on the sharding scheme, synchronization is not necessary; and relaxing it can allow some nodes to proceed earlier than others. For instance, in a sharding scheme where CompNode$_1$ holds only the first $k$ tokens, because the first $k$ logits do not depend at all on tokens $k+1,\ldots,n$ in a unidirectional model, then CompNode$_1$ can proceed through all its forward passes without waiting for \textit{any} information from other nodes. Future work could analyze the tradeoff between such synchronization relaxations, which are not possible with schemes like $(c,\delta)$-sharding, and token security.

\section{Collusion Resistance}
\label{appendix:collusion}

How many nodes in Cascade must share information to break the security of the scheme? The acceptable level of collusion depends on the sharding scheme and the desired hidden state security. Formally, let $\mathcal{N}$ be the set of all $\alpha+\beta^2$ nodes. For each node $\mathbf{n} \in \mathcal{N}$, denote $t(\mathbf{n})$ as the set of (sequence dimension) indices of hidden states that $\mathbf{n}$ directly has access to. This is explicitly
$$t(\text{CompNode}_i) = R_i, \quad t(\text{AttnNode}_{jk}) = S_j \cup T_k$$
because CompNode$_i$ directly receives hidden states with indices in $R_i$, and AttnNode$_{jk}$ directly receives $Q,K,V$-projections of hidden states with indices in $S_j$ and $T_k$.

When nodes share information, they gain access to the \textit{union} of their shards. Suppose a subset $\mathcal{S} \subseteq \mathcal{N}$ of nodes jointly colludes. Then they each gain access to hidden states with indices in the set $\cup_{\mathbf{n} \in \mathcal{S}} t(\mathbf{n})$. This can be used to reveal a significant portion of the prompt if $\mathcal{S}$ contains many nodes, so in general, Cascade is not fully collusion-resistant. For example, if all $\alpha$ CompNodes collude, then they gain access to all hidden states with indices in $\cup_{i=1}^n t(\text{CompNode}_i) = \cup_{i=1}^n R_i = [N]$, so the entire prompt is revealed by reversing layer $0$ hidden states into tokens.

Collusion also raises security concerns in vocab-matching. Recall the generalized vocab-matching attack from \cref{sec:walkthrough} can be used to break security if gaps between non-consecutive hidden states in each shard are not sufficiently large. From our prior discussion in \cref{subsubsec:security_layer_>_0} and \cref{thm:compnode-full}, we know that non-consecutive hidden states held by a node must be $> \rho$ positions apart to prevent the attack, where $\rho$ is the vocab-matching threshold. Thus, it may be the case that much fewer than $\alpha$ CompNodes need to collude to break security: even though the union set of their shards may be much smaller than $[N]$ (i.e. reveal little of the actual prompt), it could contain many pairs of non-consecutive hidden states $\leq \rho$ positions apart, and the vocab-matching attack would reverse many unknown intermediate hidden states. This means the choice of vocab-matching threshold also significantly affects the collusion limit.

Finally, we note a potential solution to security concerns from collusion: further splitting of shards assigned to nodes, akin to $m$-splitting of AttnNodes in \cref{subsubsec:security_other}. By decreasing the number of hidden states each individual node has access to, we increase the number of nodes that need to collude to break security. For instance, in the extreme case where each CompNode has access to $1$ token and the prompt contains sufficiently many filler words, a large number of CompNodes could collude and still be unable to extract private user data from the prompt. This is also resistant to the vocab-matching attack as long as CompNodes with tokens within $\rho$ positions of each other do not collude.

\end{document}